%% file: icml_2025.tex
\documentclass{article}

\usepackage{microtype}
\usepackage{graphicx}
\usepackage{caption}
\usepackage{subcaption}
\usepackage{booktabs} 
\usepackage[utf8]{inputenc} 
\usepackage[T1]{fontenc}    
\usepackage{url}            
\usepackage{booktabs}       
\usepackage{amsfonts}       
\usepackage{nicefrac}       
\usepackage{microtype}      
\usepackage{xcolor}         
\usepackage{wrapfig}
\usepackage{booktabs}
\usepackage{enumitem}
\usepackage{makecell}
\usepackage{diagbox}
\usepackage{bm}
\usepackage{dsfont}
\usepackage{multirow}
\usepackage{varwidth}
\usepackage{pifont}
\usepackage{makecell}
\usepackage{wrapfig}
\usepackage{graphicx} 
\usepackage{comment}
\usepackage{color}
\usepackage[colaction]{multicol}
\usepackage{colortbl}
\usepackage{algpseudocode} 
\usepackage{xspace}
\usepackage{rotating}
\usepackage{longtable}
\usepackage{adjustbox}
\usepackage{threeparttable}
\usepackage{listings}

\usepackage{hyperref}
\definecolor{myblue}{HTML}{bfcdf0}
\definecolor{myred}{HTML}{e07f7f}

\usepackage[preprint]{icml2024}

\usepackage{amsmath}
\usepackage{amssymb}
\usepackage{mathtools}
\usepackage{amsthm}

\usepackage[capitalize,noabbrev]{cleveref}

\theoremstyle{plain}

\theoremstyle{definition}

\theoremstyle{remark}

\usepackage[textsize=tiny]{todonotes}

\icmltitlerunning{Mixture of Hidden-Dimensions Transformer}

\begin{document}
\newcommand{\aname}{\textsc{MoHD}\xspace}
\twocolumn[
\icmltitle{
Mixture of Hidden-Dimensions Transformer
}



\icmlsetsymbol{equal}{*}

\begin{icmlauthorlist}
\icmlauthor{Yilong Chen}{iie,ucas}
\icmlauthor{Junyuan Shang}{baidu}
\icmlauthor{Zhengyu Zhang}{baidu}
\icmlauthor{Jiawei Sheng}{iie,ucas}
\icmlauthor{Tingwen Liu}{iie,ucas}
\\
\icmlauthor{Shuohuan Wang}{baidu}
\icmlauthor{Yu Sun}{baidu}
\icmlauthor{Hua Wu}{baidu}
\icmlauthor{Haifeng Wang}{baidu}
\end{icmlauthorlist}

\icmlaffiliation{iie}{Institute of Information Engineering, Chinese Academy of Sciences}
\icmlaffiliation{ucas}{School of Cyber Security, University of Chinese Academy of Sciences}
\icmlaffiliation{baidu}{Baidu Inc}

\icmlcorrespondingauthor{Tingwen Liu}{liutingwen@iie.ac.cn}

\icmlkeywords{Machine Learning, ICML}

\vskip 0.3in
]



\printAffiliationsAndNotice{} 

\begin{abstract}

Transformer models encounter challenges in scaling hidden dimensions efficiently, as uniformly increasing them inflates computational and memory costs while failing to emphasize the most relevant features for each token. For further understanding, we study hidden dimension sparsity and observe that trained Transformers utilize only a small fraction of token dimensions, revealing an "activation flow" pattern. Notably, there are shared sub-dimensions with sustained activation across multiple consecutive tokens and specialized sub-dimensions uniquely activated for each token.
To better model token-relevant sub-dimensions, we propose \aname (Mixture of Hidden Dimensions), a sparse conditional activation architecture. Particularly, \aname employs \textbf{shared sub-dimensions} for common token features and a routing mechanism to dynamically activate \textbf{specialized sub-dimensions}. To mitigate potential information loss from sparsity, we design \textbf{activation scaling} and \textbf{group fusion} mechanisms to preserve activation flow. In this way, \aname expands hidden dimensions with negligible increases in computation or parameters, enabling efficient training and inference while maintaining performance.
Evaluations across 10 NLP tasks show that \aname surpasses Vanilla Transformers in parameter efficiency and task performance. It achieves \textbf{1.7\% higher performance} with \textbf{50\% fewer activation parameters} and \textbf{3.7\% higher performance} with a \textbf{3$\times$ parameter expansion} at constant activation cost. \aname offers a new perspective for scaling the model, showcasing the potential of hidden dimension sparsity to boost efficiency.

\end{abstract}

\section{Introduction}

Large Language Models (LLMs)~\cite{claude,Gpt-4,touvronLlamaOpenFoundation2023} have demonstrated impressive performance across a wide range of natural language processing tasks. Recent research~\cite{kaplan2020scalinglawsneurallanguage} suggest that, with sufficient training data, scaling language models by increasing the number of parameters and computational resources can yield more powerful models. Nevertheless, the substantial number of parameters in LLMs often leads to significant training and inference costs. Ideally, we seek flexible model architectures~\cite{jiang2024mixtralexperts, cai2024flextronmanyinoneflexiblelarge,cai2024surveymixtureexperts}  that enable parameter scaling while maintaining computational efficiency. Specifically, the parameters in Transformers' matrices are defined by the hidden and intermediate dimensions. Some studies~\cite{qiu-etal-2024-unlocking,liu2024trainingfreeactivationsparsitylarge} observe the sparsity of intermediate dimension activations and leverage it to design adaptive networks (e.g., MoE~\cite{cai2024surveymixtureexperts,dai2024deepseekmoeultimateexpertspecialization,xue2024openmoeearlyeffortopen}) for parameter scaling or use pruning~\cite{xiaShearedLLaMAAccelerating2023,chenLoRAShearEfficientLarge2023,maLLMPrunerStructuralPruning2023} and local activation mechanisms~\cite{dejiavu} to reduce computational costs.

\begin{figure}[t]  
\centering  
\includegraphics[width=8.2cm]{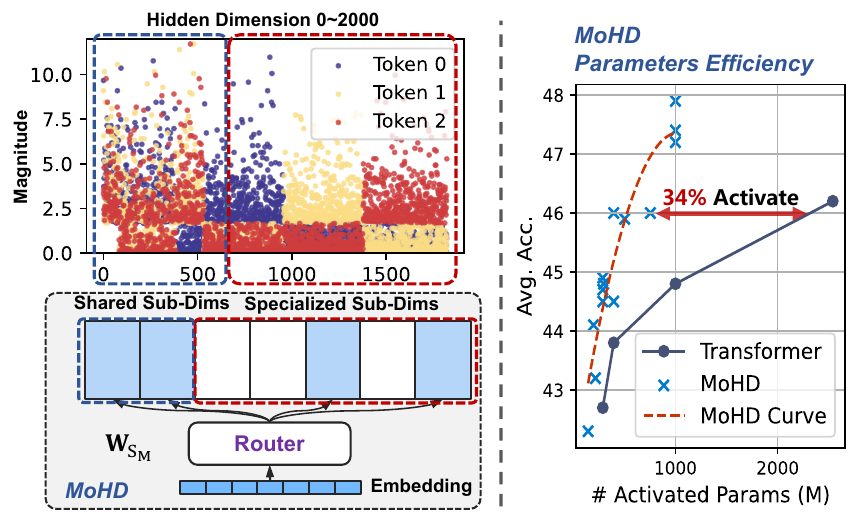}
\caption{We observe highly activated dimensions in Transformer hidden states, some shared across multiple tokens and others specific to individual tokens. Inspired by this, we propose the \aname architecture, which employs mixed activation of shared and specialized sub-dimensions. Compared to Transformers, \aname demonstrates significantly higher parameter efficiency.
}
\vspace{-2mm}
\label{fig:intro_motivation}
\end{figure}

\begin{figure}[t]  
\centering  
\includegraphics[width=8cm]{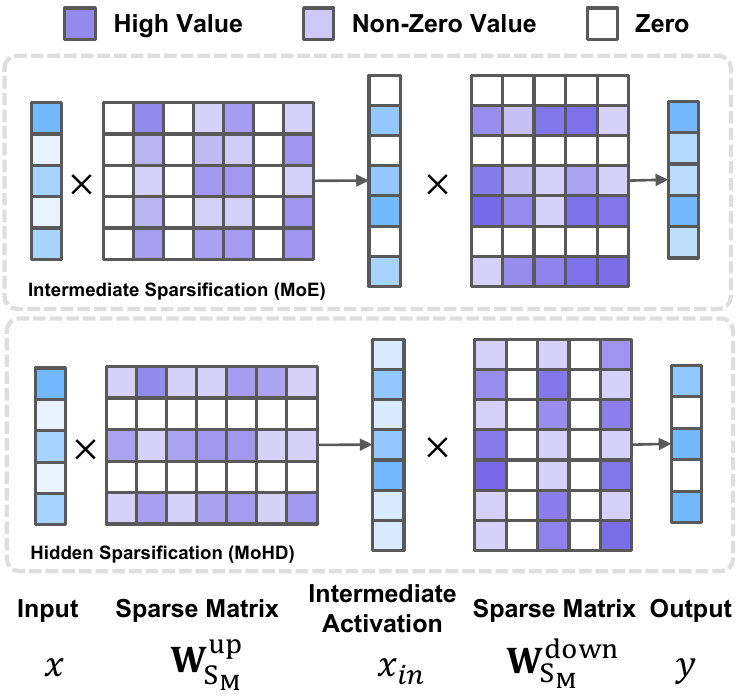}
\vspace{-2mm}
\caption{Using FFN as an example, the traditional method (MoE) exploits the sparsity of the intermediate dimension. Our method (\aname) selectively activates only a subset of hidden dimension parameters across all matrices to enhance efficiency. 
}
\vspace{-4mm}
\label{fig:motivation}
\end{figure}

While elastic scaling of the intermediate dimension is well-studied, scaling hidden dimension with controllable computational costs remains largely unexplored, as shown in Figure~\ref{fig:motivation}. The hidden dimension, representing token embedding size, models each token in a language sequence. Expanding the hidden dimension enhances model complexity and capacity, enabling it to capture more intricate patterns. However, existing Transformers~\cite{vaswani2023attentionneed} inherently treat all token dimensions equally, which leads to substantial computational and memory overhead as the hidden dimension scales up.

Given the limited understanding of hidden dimension in LLMs, we conduct an empirical study focusing on activation magnitudes. Our findings reveal \textbf{significant sparsity in the hidden dimension}, where 50\% of dimensions account for 92.54\% of the total activation magnitude (in Figure~\ref{fig:obs} Left). Among highly activated dimensions, we observe \textbf{shared dimensions consistently activated across multiple tokens and specialized dimensions activated by individual tokens}(in Figure~\ref{fig:activate3d}). Shared dimensions likely model common features across tokens, while specialized dimensions capture higher-level semantic differences and are crucial for individual token information. This observation inspires us to design an efficient network that selectively activates shared and specialized sub-dimensions of the hidden dimension for different tokens, as shown in Figure~\ref{fig:intro_motivation}. Additionally, we observe a consistent \textbf{activation flow pattern} across model layers, where \textbf{Attention and FFN exhibit distinct functional roles} regarding hidden dimension variations (in Figure~\ref{fig:obs} Middle). Attention outputs show greater variability, while FFN outputs remain stable. This insight guides the design of separate sparsity architectures for Attention and FFN, ensuring the integrity of the activation flow after sparsification.

In this paper, we propose \textbf{\aname (Mixture of Hidden-Dimensions)}, a novel approach that significantly expands the modeling capacity of the hidden dimension through sparse, conditional activation, while keeping the number of active parameters nearly unchanged from the baseline model. Specifically, \aname introduces two types of sub-dimensions at each layer of the model’s Attention and FFN components: shared sub-dimensions that are always activated to capture common dimensional information across different tokens, and specialized sub-dimensions that are selectively activated to capture token-specific specialized dimensions. Since functional roles of Attention and FFN, we train separate routing networks for each component. 
To ensure load balancing across sub-dimensions, we apply a balancing loss to the specialized sub-dimensions. An activation scaling mechanism and a grouped fusion mechanism are introduced to mitigate information loss from dimensional downsampling and maintain efficient activation flow. With proper training, \aname can be used to scale the model’s hidden dimension without increasing the number of parameters, or to significantly reduce the active hidden dimension during inference to lower computational costs.

To demonstrate the effectiveness of \aname, we pretrain Vanilla Transformer with 355M, 495M, and 1.13B parameters following the architecture of the LLaMA~\cite{touvron2023llama} and \aname Transformer in 50\%, 75\%,  2$\times$, 3$\times$, 4$\times$ settings. We evaluated these models on benchmark tasks spanning 11 different natural language processing challenges, demonstrating the advantages of the \aname architecture. Experimental results show that \aname consistently outperforms Transformer models with the same number of activated parameters across all model sizes. We found that MoHD effectively reduces activation redundancy in the model while delivering significant performance gains. In the compression setting, MoHD reduces activation parameters by 50\% while retaining 99\% of the original performance. In the expansion setting, MoHD maintains the same activation parameter count while expanding the hidden dimensions to 4× the original size, achieving up to an 8.37\% relative performance improvement. Notably, \aname-355M significantly outperformed LLaMA2-355M and even achieved performance comparable to LLaMA2-1.13B, while reducing activation parameter to LLaMA's 28.9\%. To further investigate the impact of increasing the hidden dimension, we conducted an in-depth exploration of \aname's routing mechanism and performed detailed ablation studies on sub-dimension specialization. Overall, \aname is the first method to introduce sparse mixture activation for expanding the hidden dimensions of LLMs, offering a novel perspective on designing more efficient multi-dimensional model architectures.

\section{Definition} \label{preli}

In this Section, we define the activation sparsity present in the hidden dimension of LLMs and use this to formulate sparsely activated FFN and Attention mechanisms.

Let \( X \in \mathbb{R}^{n \times d} \) denote the embeddings of $n$ tokens, and \( x \in \mathbb{R}^{1 \times d} \) represent the embedding of a single input token. The activation sparsity $\delta$ of a hidden state \( x \) is defined as the proportion of zero-valued entries within the vector~\cite{liu2024trainingfreeactivationsparsitylarge}. We then define a function \( S: d \rightarrow \delta d \) that selectively activates a subset of dimensions in \( x \). The sparsely activated representation is denoted as \( x_{s} = S(x, \delta) \), where \( x_{s} \in \mathbb{R}^{1 \times \delta d} \), representing the selective activation of \( \theta \)-proportion of the dimensions in \( x \). 

\subsection{Hidden Dimension Sparsity}

Considering the model's semantic modeling in Euclidean space, we define the magnitude $m_i$ of each dimension $i$ as the square of its activation value:
\begin{equation}
\small
m_i = x_i^2, \quad \mathbf{m} = x \odot x,
\label{eq:maginitued}
\end{equation} 
We define hidden dimension sparsity as:  
\begin{equation}
\small
\text{Sparsity} = \frac{1}{d} \sum_{i=1}^d \mathbf{1}(x_i < \epsilon),
\label{eq:Sparsity}
\end{equation} 
where \( d \) is the total number of hidden dimensions, \( x_i \) represents the squared activation value of the \( i \)-th dimension, and \( \epsilon \) is a small threshold used to identify near-zero activation values. The indicator function \( \mathbf{1}(x_i < \epsilon) \) is equal to 1 if the activation value is below the threshold and 0 otherwise.

\subsection{Hidden Sparsified FFN}

Define $\rm W^\text{up},\rm W^\text{gate} \in \mathbb{R}^{d \times d'}, \rm W^\text{down}  \in \mathbb{R}^{d' \times d}$ as the up, gate, down matrix in one FFN block, where $d'$ is the intermediate size. In this context, the \(i\)-th row of the up, gate matrix is defined as \( \rm W^{\text{up}}_i,\rm W^{\text{gate}}_i \in \mathbb{R}^{1 \times d'} \), and the \(i\)-th column of the down matrix is defined as \(\rm W^{\text{down}}_i \in \mathbb{R}^{d' \times 1} \). Specifically, the sparsely activated hidden state \( x_s \) under activation sparsity \( \delta \)  only activates a subset of rows in the up, gate matrix and a corresponding subset of columns in the down matrix, denoted as \( S_M \subseteq [d] \). Thus, the sparsified FFN computation can be described as follows:
\begin{equation}
\small
\text{FFN}_{S_H }(x_s) = \rm W_{S_M }^\text{down} \left( \sigma\left( x_{s} \rm W_{S_M }^\text{up} \odot x_{s} \rm  W_{S_M }^\text{gate} \right) \right),
\label{eq:smlp}
\end{equation}
where $\sigma$ is the activation function. $\odot$ is the element-wise production. Due to the sparsification of the hidden state, the up and gate matrices share the same activation subset. To ensure the output remains sparsified, the down matrix is also sparsified, though its activation subset can differ from that of the up and gate matrix.

\subsection{Hidden Sparsified Attention}

For a $h$-head Multi-Head-Attention (MHA), we define $\rm W_i^\text{Q}, \rm W_i^\text{K}, \rm W_i^\text{V} \in \mathbb{R}^{d \times d_h},\rm W_i^\text{O}  \in \mathbb{R}^{d_h \times d}$ as key, query, value and output projections for the $i$-th head, where $d_h$ denotes as the head dim,  \( i \subseteq [h] \). 
With sparsely activated hidden state \( x_s \), a small parameter subset \( S_A \) represents a sparsely activated selection of rows from $\rm W_i^\text{Q},\rm W_i^\text{K},\rm W_i^\text{V}$ and columns from $\rm W_i^\text{O}$.
\begin{equation}
\small
\text{MHA}_{S_A}(x_s) = \sum_{i=1}^{h} \text{Head}_i \rm W_{i,S_A}^\text{O},
\label{eq:sattn}
\end{equation}
\begin{equation}
\small
\text{Head}_i = \sigma \left( \left( x_s \rm W_{i,S_H}^\text{Q} (x_s \rm W_{i,S_H}^\text{K})^\top \right) \frac{1}{\sqrt{d_h}} \right) x_s \rm W_{i,S_H}^\text{V},
\label{eq:sattn}
\end{equation}
where $\sigma$ is the softmax function. Since \( x \) is sparse in the hidden dimension, we can find an approximation \( x_s \) of \( x \), such that, under the activation of the corresponding subset of parameters \( S_H \), the outputs of the sparsified FFN and sparsified attention closely approximate the outputs of the dense model.

\section{Observation}\label{obs:sub-dimensions}

\begin{figure*}[t]
\begin{minipage}[c]{\textwidth}
\centering
  \includegraphics[width=0.98\textwidth]{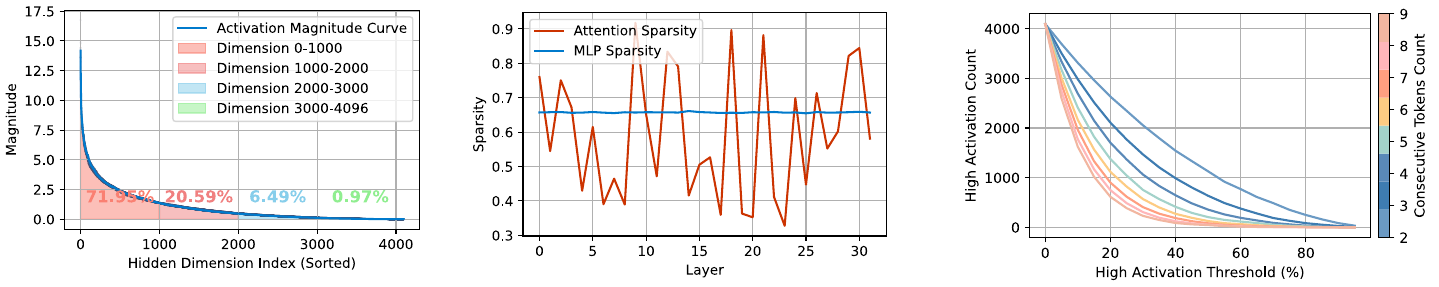}
  \caption{Visualization of the hidden dimension activation pattern in LLaMA2-7B. Left: Activation magnitudes sorted in descending order with the percentage representing the cumulative activation sum. Middle: Sparsity of hidden dimension activations in Attention and FFN outputs across layers. Right: Number of shared activation dimensions at varying activation magnitude thresholds, with curves showing the count for consecutive token ranges from 2 (blue) to 20 (red).}\label{fig:obs}
\end{minipage}
\end{figure*}

In this section, we present several key findings that serve as the foundation for the design of the \aname approach. In Section~\ref{obs.sparsity}, we observe the long-tail effect of hidden dimension activation values and define activation sparsity accordingly. We analyze the sparsity distribution and differences between attention and FFN across different layers. In Section~\ref{obs.flow}, we analyze activation flow in Transformers, highlighting compression patterns, stabilization by residuals and normalization, and functional layer differences.
In Section~\ref{obs.con_activate}, we further identify the existence of shared continuous high activation behaviors and unique discrete high activation behaviors across tokens. Finally, we analyze these phenomena and propose motivations for designing feasible hidden dimension sparsification methods.

\subsection{Sparsity in Tokens' Hidden Dimension}\label{obs.sparsity}

For a more comprehensive understanding, we observe the activation magnitudes of 4096 hidden dimensions in LLaMA2-7B. As shown in the left panel of Figure~\ref{fig:obs}, we visualize the relationship between dimension magnitudes and reordered dimension indices based on magnitude size.

Similar to previous observations~\cite{liu2024trainingfreeactivationsparsitylarge}, the activation of hidden dimensions exhibits a long-tail sparsity phenomenon. For instance, in the input Attention activations of LLaMA2-7B’s 16th layer, the cumulative magnitude of the top 1000 dimensions accounts for 71.96\% of the total magnitude. In contrast, most dimensions have low activation values, indicating that the model does not utilize information from the majority of hidden dimensions, leading to substantial sparsity in activations. 

We also visualized the sparsity of activations in the input and output of Attention and FFN components. Our observations reveal \textbf{significant differences in the magnitude of hidden activations across positions.} Attention exhibits higher activation magnitudes, while FFN activations are comparatively lower. At the input stage, activation magnitudes are relatively high (median > 1), whereas at the output stage, activation magnitudes drop significantly (median < 0.5). The sparsity of hidden dimensions in the input components is consistent across different modules, likely due to the influence of residual connections. However, \textbf{at the output stage, the sparsity patterns of Attention and FFN differ markedly.} As shown in the middle panel of Figure~\ref{fig:obs}, Attention demonstrates significant fluctuations in sparsity, with alternating high and low sparsity distributions. In contrast, FFN sparsity remains relatively stable. These differences highlight the distinct functional roles and information processing characteristics of Attention and FFN, prompting us to consider differentiated activation designs for these components.

\subsection{Activation Flow in Transformer} \label{obs.flow}
\begin{figure}[t]  
\centering  
\includegraphics[width=8cm]{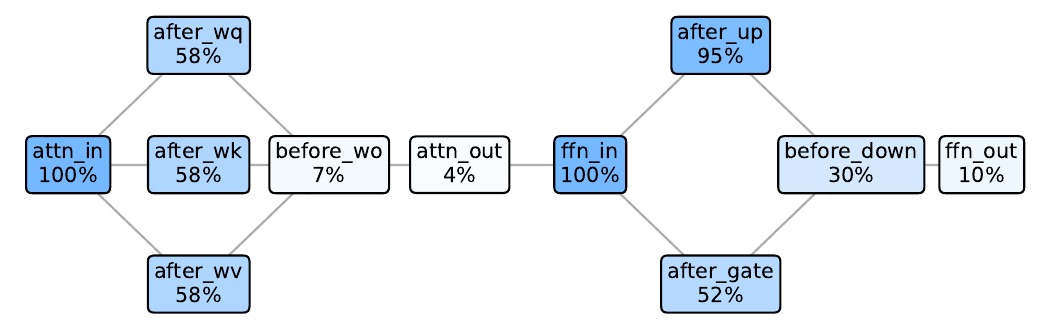}
\caption{Visualization of activation magnitude in LLaMA2-7B layer 30. In the Transformer, multiple layers show a consistent pattern of activation flow.
}
\label{fig:activation_flow}
\end{figure}
We also investigate the variations in activation magnitudes within a single Transformer block, as illustrated in Figure~\ref{fig:activation_flow}. \textbf{Consistent activation flow patterns were observed across different Transformer blocks.} The Attention module compresses input activations normalized to 100\% through projections (\(W_Q\), \(W_K\), \(W_V\)) and weighted averaging, reducing activation magnitudes to 6.7\% at \(W_O\). This highlights its ability to suppress irrelevant information through weighted aggregation, while also \textbf{showcasing significant functional differences between layers}, as the output activation magnitudes vary to accommodate layer-specific roles. In contrast,\textbf{the FFN module demonstrates stable activation patterns}, with compression arising from high-dimensional projections, nonlinearity that sparsifies activations, and dimensionality reduction through linear weighted summation, collectively reducing activation magnitudes.



\textbf{Residual connections play a crucial role in regulating activation magnitude changes.} In the outputs of the Attention and FFN modules, residual connections directly add the input back to the output, partially restoring the compressed activation magnitudes. Layer Normalization further balances and constrains activation magnitudes, stabilizing the numerical distribution and suppressing excessively high or low activation values, thereby enhancing the training stability of the Transformer. However, this normalization also smooths activation change patterns, potentially diminishing the prominence of contextual information and further compressing attenuated activation magnitudes, particularly exacerbating the instability of Attention outputs.

\subsection{Continuous High Activation}\label{obs.con_activate}  
\begin{figure}[t]  
\centering  
\includegraphics[width=5cm]{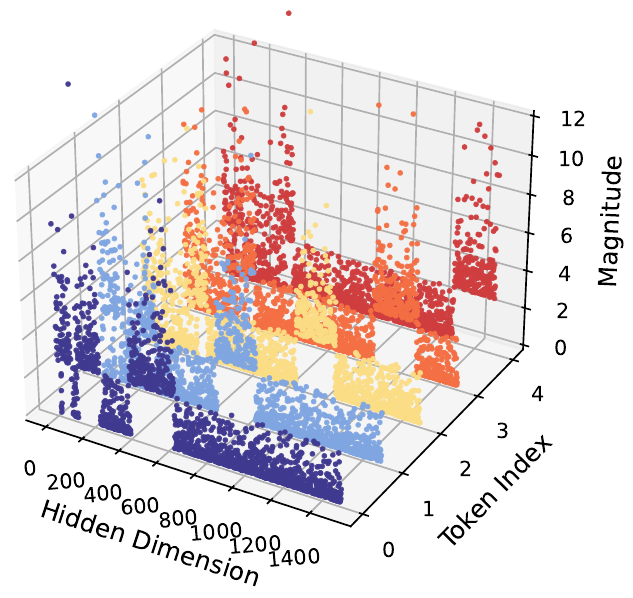}
\caption{Activation patterns of 4,096 hidden dimensions clustered and reordered across five tokens in LLaMA2-7B 16th layer. Around 400 dimensions show consistent high activation, modeling token similarity, while ~200 dimensions per token exhibit unique high activation, highlighting differences..}
\label{fig:activate3d}
\end{figure}
We further investigate the temporal correlation of activation sparsity by observing high activation values across different tokens and analyzing the indices that are repeatedly activated by multiple tokens.\textbf{ A clear correlation in activations is observed over consecutive tokens.} Figure~\ref{fig:obs} Right shows the number of commonly highly activated dimensions across 2 to 9 consecutive tokens, with the x-axis representing the threshold for defining high activation. When using the top 20\% of activation values as the threshold, 2672 dimensions are commonly activated across 2 consecutive tokens, and 673 dimensions remain commonly activated across 9 consecutive tokens. 

Figure~\ref{fig:activate3d} further illustrates the correlated activation patterns over 5 tokens, where the 4096 hidden dimensions are clustered and reordered based on their activation patterns. Approximately 400 dimensions are commonly highly activated across all 5 tokens, while about 200 dimensions are uniquely highly activated within each token. This indicates that \textbf{each token's activations contain shared sub-dimensions that are commonly activated and token-specific sub-dimensions that are independently activated.} Shared high activations model the similarity information shared across tokens in hidden dimensions, while specialized unique activations capture differences. These observations inspired the shared-specialized activation mechanism in the subsequent design of MoHD.

\begin{figure*}[t]
\begin{minipage}[c]{\textwidth}
\centering
  \includegraphics[width=\textwidth]{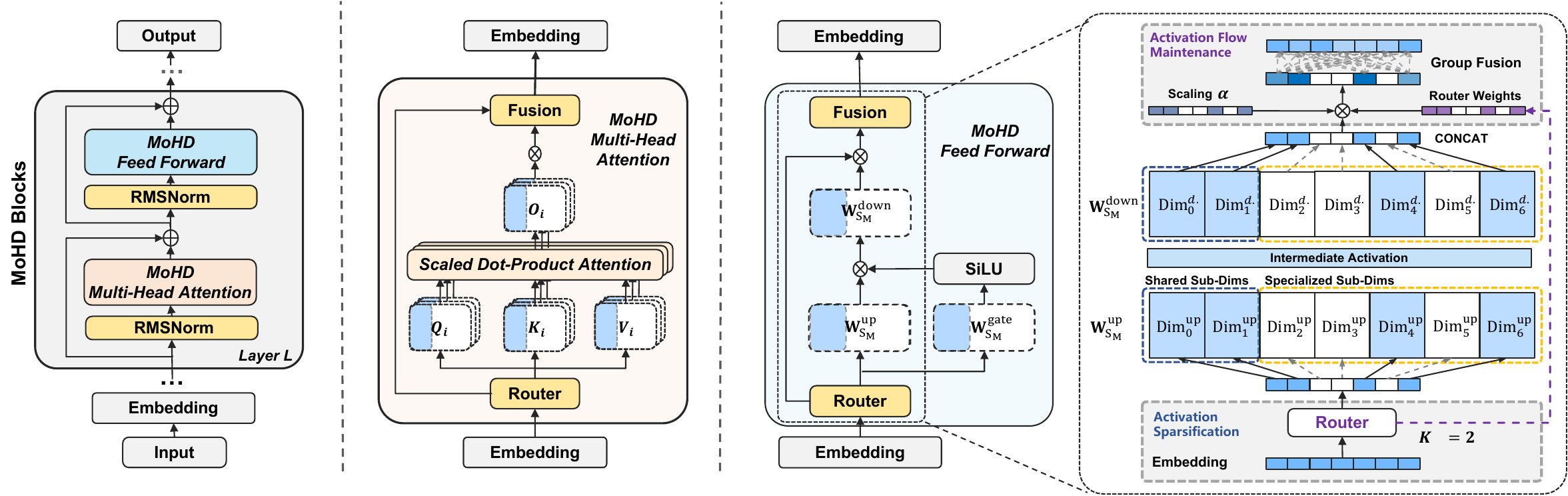}
  \caption{An illustration of \aname. A single \aname Block follows the same structure as a LLaMA Block, consisting of two key components: \aname Attention and \aname FFN, both equipped with pre-norm and residual connections. In \aname Attention and \aname FFN, we selectively activate the matrices in each component according to the dimensions chosen by the Router. As illustrated on the right, the \aname router selects certain shared dimensions along with a few sparsely activated dimensions for each input token. The outputs generated from these sparsely activated matrices are then weighted and concatenated based on the router's weights, before being mapped back to their original dimensions using the group fusion matrix.}\label{fig:main}
\end{minipage}
\end{figure*}

\section{Mixture of Hidden Dimensions (\aname)}\label{sec:method}

In this Section, we propose the Mixture of Hidden Dimensions (\aname) architecture to expand the hidden dimension of the model without increasing the number of activated parameters. The full workflow is illustrated in Figure~\ref{fig:main}. 
Inspired by the observations on LLM's hidden dimension activation phenomena discussed in Section~\ref{obs.con_activate}, we introduce the Shared and Specialized Sub-Dimension Mixed Activation mechanism in Section~\ref{method:experts}. Furthermore, we present the implementation of the sparsified components, as defined in Section~\ref{preli}, applied to both the Attention and FFN blocks. In Section~\ref{method:flow}, we address the issue of information degradation and how we mitigate it using activation scaling and grouped fusion mechanisms. In Section~\ref{method:loss}, we explore the design of a balancing loss to enhance diversity. Finally, in Section~\ref{method:impl}, we detail the optimized implementation of \aname.

\subsection{Mixture of Sub-Dimensions Activation}
\label{method:experts}

As defined in Section~\ref{preli},  \( X \in \mathbb{R}^{n \times d} \) denote the embeddings of $n$ tokens, and \( x \in \mathbb{R}^{1 \times d} \) represent the embedding of a single input token. Under a specific activation sparsity $\delta$, we selectively activate a subset $ S \subseteq [d]$ of parameters of a matrix $\rm W \in \mathbb{R}^{d \times d'}$. \textit{Inspired by the hidden dimension sparsity observed in Section~\ref{obs.sparsity}, we can selectively utilize a subset of each token's hidden dimensions.} Therefore, we construct \( M \) sub-dimensions by slicing the weight matrix \( \rm W \) along the hidden dimension, where each sub-dimension has a dimension of \( d_e \). Specifically, \( \rm W = [\rm W[0:d_e], \rm W[d_e:2d_e], ..., \rm W[(N-1)d_e:N d_e]] \), with \( E d_e = d \). Here, \( \text{Dim}_1 : \rm W[d_e:2d_e] \) represents the sub-parameters of \(\rm W \) from \( d_e \) to \( 2d_e \). \textbf{\aname leverages a dynamic routing mechanism to select a subset of sub-dimensions for each token, which allows the model to avoid involving the entire hidden dimension. }

 Define the routing gate \( g,x_s =\text{Gate}(x,\delta, N) \) determines the top-\( K \) sub-dimensions to activate from the \( N \) sub-dimensions based on sparsity $\delta$,  assigns each activated $i$-th sub-dimension a weight \( g_i \): 
\begin{equation}
\small
s_{i}=\operatorname{Softmax}_i\left(x^T \phi_i^l\right),
\label{eq:sattn}
\end{equation}
\begin{equation}
\small
g_{i}= \begin{cases}s_{i,}, 
& s_{i} \in \operatorname{Topk}\left(\left\{s_{j} \mid  1 \leqslant j \leqslant  N\right\},  N\right), \\ 0, 
& \text { otherwise, }\end{cases}
\label{eq:sattn}
\end{equation}
where $s_{i, t}$ denotes the token-to-sub-dimension score, $\operatorname{Topk}(\cdot, K)$ denotes the set comprising $K$ highest affinity scores among those calculated for the input token and all sub-dimensions, and $\phi_i^l$ is the centroid of the $i$-th sub-dimension in the $l$-th layer. All sub-dimensions are assigned routing weights, allowing the routing mechanism to learn to selectively amplify or suppress the representations of shared sub-dimensions during the optimization process. Finally, the outputs $y_s$ from all activated sub-dimensions are concatenated and weighted, resulting in a final output of dimension \( d \), which matches the hidden dimension.
\begin{equation}
\small
y_s = \big\|_{i=1}^{N} g_i \, \text{Dim}_i(x_s)
=\rm W_{S} x_s.
\label{eq:concat}
\end{equation}
We use the notation \(\big\|_{i=1}^{N} g_i \, \text{Dim}_i(x)\) to denote the concatenation of the terms \(g_i \, \text{Dim}_i(x_s)\) for \(i = 1, \dots, N\). Due to the high sparsity of the gate, only a small subset of dimensions is assigned non-zero weights, while most dimensions remain zero. In practice, based on the gate’s selection, we can sparsify \( x \) and \( \rm W \), and finally get output \( y_s \in \mathbb{R}^d\), meaning the number of activated parameters in \( \rm W_S \) is reduced to \( \delta \) of the original.

\subsection{Activation Flow Maintenance} \label{method:flow}

\begin{figure}[t]  
\centering  
\includegraphics[width=8cm]{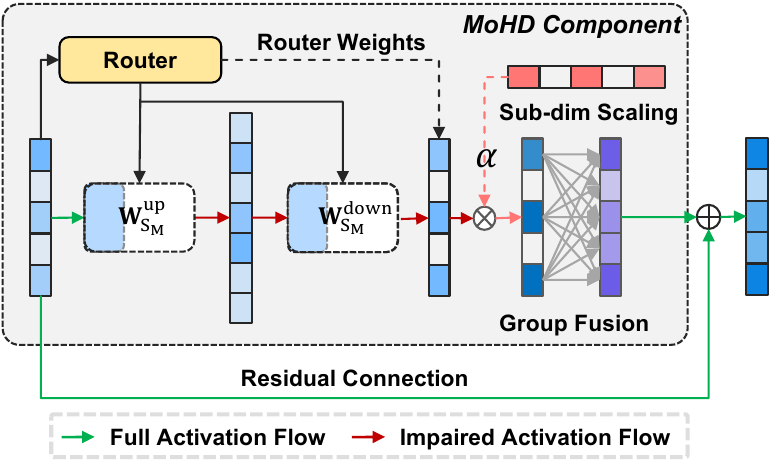}
\caption{An illustration of \aname's Activation Flow Maintenance. Sparse activations represent only a small subset of the total hidden dimensions, leading to inevitable information loss. The scaling vector scales the activation information to match the original level and grouping fusion mechanism fuses it back into the original hidden dimensions.}
\label{fig:flow}
\end{figure}

In Section \ref{method:experts}, we sparsely activate a subset of parameters, resulting in a sparsified hidden-dimension output $y_s$, which substantially reduces computational costs. However, a challenge arises due to the router assigning softmax-normalized weights to different dimensions. This may lead to a few dimensions receiving disproportionately high weights, while information within many other dimensions may be neglected due to their low assigned weights. Additionally, because we concatenate the final output in parallel, any weight below 1 suppresses information within that sub-dimension without compensating for this loss through other sub-dimension. Unlike Mixture of Experts methods~\cite{zhou2022mixture,jiang2024mixtral}, \aname directly use concatenation for integration, which can lead to information degradation. \textit{Motivated by the activation flow observed in Section~\ref{obs.flow}, we employ the following strategies to maintain robust activation flow while preserving sparse activation:} \textbf{sub-dimension scaling}, \textbf{grouped dimension fusion}, and \textbf{residual connections}.

To address the suppression of sub-dimension activations caused by the softmax weight normalization, we introduce a scaling factor to ensure that the sum of activation weights across all dimensions remains consistent with the input. We define the scaling factor \( \alpha = \sum g_i N\), which ensures that the activated dimensions retain their proportional influence. To address the information loss from the sparse output \( y_s \), we employ a fusion mapping layer that projects \( y_s \) from its activated sub-dimensions back to the original dimension \( d \). To reduce computational overhead, we introduce a Monarch matrix~\cite{daoMonarchExpressiveStructured2022a, chen-etal-2024-lemon} to perform grouped fusion mapping. Given a receptive field \( r \), we define the mapping matrix $\rm M \in \mathbb{R}^{d \times d}, \rm M=\sum_{i=1}^{d / r} \sum_{j=1}^{d / r} m_{i, j}$ as follows:
\begin{equation}
\small
\rm M y_s = \left[\begin{array}{ccc}
m_{1,1} & \cdots & m_{1, d / r} \\
\vdots & \ddots & \vdots \\
m_{d / r, 1} & \cdots & m_{d / r, d / r}
\end{array}\right] \otimes y_s,
\end{equation}
where the Monarch matrix \( M \) enables efficient grouping and transformation, thereby reconstructing the information across the original hidden dimensions while keeping computations tractable.
In summary, the forwarding process for a single \aname module can be formally represented as follows:
\begin{equation}
\small
y = \rm M \alpha \big\|_{i=1}^{N} g_i \, \text{Dim}_i(x_s), ~~~g,x_s = \text{Gate}(x,\delta, N).
\label{eq:mohmatrix}
\end{equation}

\subsection{Mixed Activated Sub-Dimensions}

As discussed in Section~\ref{obs.con_activate}, \textit{a portion of the hidden dimensions is always activated, potentially containing important shared features, while another portion is selectively activated, likely representing token-specific differentiated features.} Therefore, we designed two types of sub-dimensions in \aname:  \textbf{Shared Sub-Dimensions} and \textbf{Specialized Sub-Dimensions}. Shared sub-dimensions are always activated by the routing mechanism, whereas Specialized sub-dimensions are selectively activated based on the routing decisions. Define $\varphi$ as the percentage of shared sub-dimensions activation rate, the routing gate \( \text{Gate}(x,\delta, \varphi) \) determines the top-\( K \) sub-dimensions to activate from the \( N \) sub-dimensions based on sparsity, and assigns each activated $i$-th sub-dimension a weight \( g_i \): 
\begin{equation}
\small
s_{i}=\operatorname{Softmax}_i\left(x^T \phi_i^l\right),
\label{eq:sattn}
\end{equation}
\begin{equation}
\small
g_{i}= \begin{cases}s_{i,}, 
& s_{i} \in \left\{s_{j} \mid 1 \leqslant j \leqslant  \varphi N\right\}, \\s_{i},
& s_{i} \in \operatorname{Topk}\left(\left\{s_{j} \mid  \varphi N \leqslant j \leqslant  N\right\}, (\delta - \varphi ) N\right), \\ 0, 
& \text { otherwise, }\end{cases}
\label{eq:sattn}
\end{equation}
where $s_{i, t}$ denotes the token-to-sub-dimension score, $\operatorname{Topk}(\cdot, K)$ denotes the set comprising $K$ highest affinity scores among those calculated for the input token and all sub-dimensions, and $\phi_i^l$ is the centroid of the $i$-th sub-dimension in the $l$-th layer. Under this routing mechanism, \textbf{all tokens are consistently activated in the dimensions associated with Shared Sub-Dimensions, which consolidate and capture common information.} This, in turn, \textbf{encourages the differentiation and diversification of Specialized Sub-Dimensions.} However, all sub-dimensions are assigned routing weights, allowing the routing mechanism to learn to selectively amplify or suppress the representations of all sub-dimensions during the optimization.

\subsection{Sub-Dimension Load Balance} \label{method:loss}
Research on conditional computation~\cite{zhou2022mixture,jiang2024mixtralexperts,dai2024deepseekmoeultimateexpertspecialization} has shown that automatically learned routing strategies can often lead to load imbalance issues, where the model tends to select only a few sub-dimensions, leaving others underutilized and insufficiently trained. \textit{To distribute tokens more evenly among different sub-dimensions and smooth out the router score distribution},  we incorporate Sub-Dimension Load Balance Loss~\cite{dai2024deepseekmoeultimateexpertspecialization}: Define \(\beta\) is a scaling factor and \( \mathbb{1}_{\{ \text{argmax}(g_s) = i \}} \) is an indicator function that returns 1 if the \( i \)-th sub-dimension has the highest gating score for the \( s \)-th sequence position and 0 otherwise. 
\begin{equation}
\small
 \mathbb{L}_{\text{B}} = \beta \sum_{i=1}^{N} \frac{g_i}{\sum_{j=1}^{N} g_j} \cdot \frac{\sum_{s \in S} \mathbb{1}_{\{ \text{argmax}(g_s) = i \}}}{M}.
\end{equation}
The term \(\frac{g_i}{\sum_{j=1}^{N} g_j}\) represents the normalized gating score for sub-dimension \( i \), ensuring that the contributions of each sub-dimension are proportional to their selection frequency. The auxiliary loss thus encourages the gating mechanism to distribute \textbf{the assignments more evenly across sub-dimensions} by penalizing imbalances, ultimately leading to improved model performance and efficiency.

\subsection{Implementation}\label{method:impl}

In Sections~\ref{obs.sparsity} and~\ref{obs.flow}, we observed activation differences across components in various layers, prompting us to design separate routing mechanisms for the Attention and FFN components.  Specifically, in one Transformer Block, $\text{Gate}_{\text{attn}}(x,\delta, N, \varphi)$ and $\text{Gate}_{\text{ffn}}(x,\delta, N, \varphi)$ producing scores that determine the activation of dimension-specific sub-dimensions for the output: 
\begin{equation}
\small
a,x_s = \text{Gate}_{\text{attn}}(x,\delta, N,\varphi), m,x_s = \text{Gate}_{\text{ffn}}(x,\delta, N, \varphi). \nonumber
\label{eq:implegate}
\end{equation}
In practice, different components may employ distinct sparsification settings. However, for simplicity, we use the same notation throughout this section to represent these settings in a unified manner. Based on the scores from the Router, \aname applies synchronized sparsification to the hidden dimensions of all up-projection and down-projection matrices, as well as the input $x$.  From Equation~\ref{eq:mohmatrix}, we transform $\rm W^\text{Q}, \rm W^\text{K}, \rm W^\text{V}, \rm W^\text{O}, \rm W^\text{up}, \rm W^\text{gate}, \rm W^\text{down}$ into \aname’s sub-dimensions $\big\|_{i=1}^{N} \, \text{Q}_i,\big\|_{i=1}^{N} \, \text{K}_i,\big\|_{i=1}^{N} \, \text{V}_i,\big\|_{i=1}^{N} \, \text{O}_i$  and $\big\|_{i=1}^{N} \, \text{UP}_i,\big\|_{i=1}^{N} \, \text{GATE}_ji,\big\|_{i=1}^{N} \, \text{DOWN}_i$. We substitute these into the sparsified Attention and FFN defined in Equation~\ref{eq:sattn} and~\ref{eq:smlp}, yielding outputs \( y_a \) and \( y_m \), respectively:

\begingroup
\vspace{-6mm}
\small
\begin{align}
\text{MHA}_{\text{\aname}}(x_s)  &= \rm M_a \alpha_a \sum_{i=1}^{h} \text{Head}_i \left( \big\|_{j=1}^{N} a_j \, \text{O}_j(x_s) \right), \\
\text{Head}_i &= \sigma \left( \left( x_s \left( \big\|_{j=1}^{N} a_j \, \text{Q}_j(x_s) \right) \right. \right. \nonumber \\
& \quad \left. \left. \left( x_s \left( \big\|_{j=1}^{N} a_j \, \text{K}_j(x_s) \right) \right)^\top \right) \frac{1}{\sqrt{d_h}} \right) \nonumber \\
& \quad \times x_s \left( \big\|_{j=1}^{N} a_j \, \text{V}_j(x_s) \right), 
\label{eq:mohattn}
\end{align}
\endgroup
\begingroup
\small
\begin{align}
\small
\text{FFN}_{\text{\aname}}(x_s)  &= \rm M_m \alpha_m \left( \bigg\|_{i=1}^{N} m_i\, \text{DOWN}_i(x_s) \right) \nonumber \\
&\quad \times \sigma\Bigg(
    x_s \left( \bigg\|_{i=1}^{N} m_i\, \text{UP}_i(x_s) \right) \nonumber \\
&\qquad \odot\,
    x_s \left( \bigg\|_{i=1}^{N} m_i\, \text{GATE}_i(x_s) \right)
    \Bigg).
\label{eq:mohmlp}
\end{align}
\endgroup
We construct a \textbf{\aname BLOCK} based on \aname specified MHA and FFN components. Residual connections are designed to further mitigate information loss during the specified forward pass. Following the configuration of LLAMA, we apply LayerNorm layers before the input to both MHA and FFN; however, for simplicity, these are omitted in the formal equations. This process can be formalized as follows:
\begin{equation}
\small
    \text{BLOCK}_{\text{\aname}}(x) = \text{FFN}_{\text{\aname}}(\text{MHA}_{\text{\aname}}(x)+ x) + x.
\end{equation}
 To train the model effectively, we combine cross-entropy loss \( \mathbb{L}_{\text{CE}} \) for language model pre-training and the load balance loss \( \mathbb{L}_{\text{B}} \), resulting in the final training objective:
\begin{equation}
  \mathbb{L} = \mathbb{L}_{\text{CE}} +  \mathbb{L}_{\text{B}}.
\end{equation}

\section{Experiments}

\subsection{Experimental Setup}

\paragraph{Data.} To pretrain \aname models and baseline models, we employ the RedPajama~\cite{Redpajama}, which parallels the LLaMA training data across seven domains: CommonCrawl, C4, GitHub, Wikipedia, Books, ArXiv, and Stack-Exchange. This dataset comprises a validation set with 2 million tokens, a training set containing 50 billion tokens.

\paragraph{Training.} Our experimental framework utilizes the Sheared-LLaMA codebase \cite{xiaShearedLLaMAAccelerating2023} implemented on the Composer package \cite{mosaicml2022composer}, and is executed on 8 NVIDIA A100 GPUs (80GB). The models are trained with a sequence length of 4096, employing a global batch size of 64 during the fusion phase and 256 during the continued pre-training phases.
\aname models were trained for 50000 steps (50B token budget). The learning rates were set at 3e-4 for both model parameters and router parameters. The baselines and all \aname models follow the same training setup, starting from random initialization and training on the same amount of data.

\paragraph{Evaluation.} We employed the lm-evaluation-harness \cite{eval-harness} to evaluate our models. For common sense and reading comprehension tasks, we report 0-shot accuracy results for SciQ \cite{sciqa}, PIQA \cite{piqa}, WinoGrande (WG) \cite{WinoGrande:conf/aaai/SakaguchiBBC20}, ARC Easy(ARC-E) \cite{clark2018think}, and 10-shot HellaSwag (Hella.) \cite{HellaSwag:conf/acl/ZellersHBFC19}, alongside 25-shot accuracy for ARC Challenge (ARC-C) \cite{arcChallenge:journals/corr/abs-1803-05457}. In the assessments of continued QA and text understanding, we report 0-shot accuracy for LogiQA \cite{liu2020logiqa}, 32-shot BoolQ \cite{clark2019boolq}, and 0-shot LAMBADA (Lam.) \cite{paperno2016lambada}. All reported results were calculated with the mean and stderr of multiple experiments.

\paragraph{Baseline.} Following the architecture of LLaMA2, we constructed models at three parameter scales: 355M, 495M, and 1.13B, with hidden dimensions of 1024, 1536, and 2048. At each parameter scale, we developed three variants: a standard Transformer model (LLaMA architecture) and an \aname-based model. Due to the flexibility of the \aname architecture, we can compress the model activation parameters and keep the number of parameters unchanged, or we can keep the activation parameters and expand the equivalent number of model parameters. For each \aname model scale, we experimented with five different hidden dimension scaling factors—$0.5\times$, $0.75\times$, $2\times$, $3\times$, and $4\times$—to demonstrate \aname's potential in both reducing computational cost and effectively increasing model capacity. All models were initialized with the same random seed and pre-trained on a uniform dataset of 50 billion tokens.

\input{tables/model_setting}

\subsection{Result}

\input{tables/main}
\input{tables/moh_setting}
\paragraph{Capability in Compression.}
Table~\ref{table:main-results1} demonstrates the fundamental capabilities of \aname on the 355M, 495M, and 1B versions of LLaMA2 after activating only 50\% and 75\% of the hidden dimensions. All models were trained from scratch. Results indicate that when only part of the hidden dimensions are activated, \aname maintains, and in some cases even improves, its performance. Specifically, at the 355M scale, \textbf{\aname with only 50\% activated parameters incurs an average performance loss of merely 0.4\% compared to the baseline}, highlighting the high sparsity observed in hidden dimension activation. More notably, at all model scales, \textbf{\aname with 75\% hidden dimension activation outperforms the fully activated baseline}, with average performance gains of 0.5\%, 1\%, and 1.8\% for the 355M, 495M, and 1B models, respectively. This suggests that full hidden dimension activation during training is not optimal; instead, \aname achieves higher parameter efficiency by selectively activating hidden dimensions and encouraging token-specific activation patterns.  Finally, we observe that \textbf{\aname's performance gains increase with model size}. For instance, \aname with 50\% activation demonstrates a relative improvement over the baseline of -0.4\%, +0.3\%, and +1.7\% for the 355M, 495M, and 1B models, respectively. This indicates promising potential for \aname in larger-scale models. Finally, we found that highly compressing the original hidden dimension activation (\aname 50\%) leads to a slight decrease in Commonsense metrics. However, \aname still maintains strong LM metrics under low activation settings, demonstrating the model’s robust language modeling capabilities.

\paragraph{Capability in Extension.}
Table~\ref{table:main-results1} presents the baseline capabilities of \aname on LLaMA2 models at the 355M, 495M, and 1B scales under 2$\times$, 3$\times$, and 4$\times$ hidden dimension expansion. \textbf{\aname demonstrates strong scalability}, achieving performance comparable to models with an equivalent number of effective parameters, while maintaining a substantially lower count of activated parameters. For instance, in the 355M model, \aname with 2$\times$ activation (equivalent to 710M effective parameters) surpasses the baseline performance by 2.2\%, even outperforming the LLaMA2-495M and LLaMA2-1.13B models, which have higher numbers of activated parameters. This improvement underscores \aname's effective utilization of hidden dimension sparsity, leveraging differentiated hidden dimension activation to boost performance.
In the 2$\times$ configuration, \aname achieves performance gains of 2.2\%, 0.7\%, and 3\% for the 355M, 495M, and 1.13B models, respectively. This indicates that \textbf{as model scale increases, the benefit of parameter expansion with \aname grows proportionally.} Experimental results further show that parameter expansion with \aname yields significant improvements across multiple tasks, including natural language modeling (LAMBADA), reading comprehension (SciQ, ARC-E, ARC-C, HellaSwag).However, the performance improvement was relatively small on the LogiQA and MMLU datasets. Finally, \textbf{as we increased the overall parameter of \aname, we observed a performance improvement}, with optimal results often achieved when the hidden dimension was tripled. For example, \aname×3 with 1.48B parameters showed a 2.2\% improvement over the baseline and a 1.5\% improvement over \aname×2 with 989M parameters. The routing mechanism in \aname effectively increases the equivalent hidden dimension, enabling significant performance gains, although a performance ceiling exists under the same activation count. Overall, \aname showcases a structural advantage for building large-scale models.

\begin{figure}[t]
\centering
\begin{subfigure}{0.45\textwidth}
    \centering
    \includegraphics[width=\linewidth]{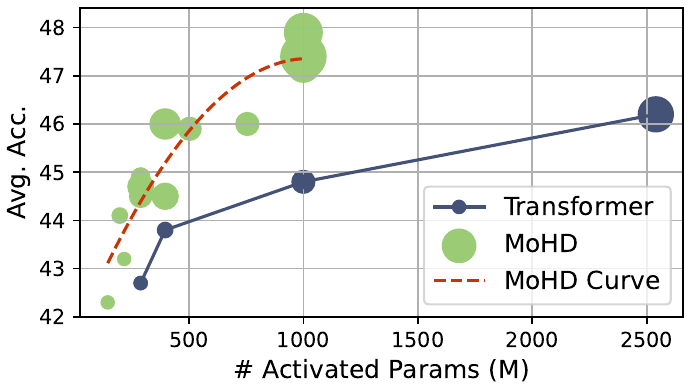}
    \caption{Average Score with Activated Parameters. Point size represents the model's All Parameters.}
    \label{fig:acti-p.w.avg}
\end{subfigure}
\hspace{0.05cm}
\begin{subfigure}{0.45\textwidth}
    \centering
    \includegraphics[width=\linewidth]{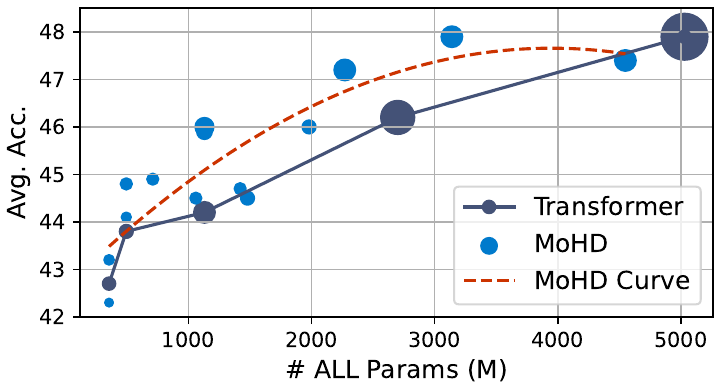}
    \caption{Average score with All Parameters. Point size represents the model's Activated Parameters.}
    \label{fig:all-p.w.avg}
\end{subfigure}
\caption{A comparison of average test accuracy on downstream tasks between \aname and baseline models ...}
\label{ablation study figures}
\end{figure}

\paragraph{Parameters Efficiency.}In Figure~\ref{fig:acti-p.w.avg}, we analyze the relationship between model performance and model size from the perspectives of activation parameters and total parameters. When examining model performance under equal activation parameter conditions, we found that \textbf{\aname achieves exceptionally high parameter efficiency}. Compared to the baseline, various \aname models at the 400M and 1B scales achieved absolute improvements of approximately 2.2\% and 3\%, respectively. At comparable performance levels, \aname often requires activation of less than 50\% of the original model parameters. As the activation parameter count increases, \textbf{the performance gain of \aname over the baseline grows, underscoring its advantages at larger parameter scales}. In Figure~\ref{fig:all-p.w.avg}, we further analyze the performance of \aname and the baseline as total parameters increase. At smaller model scales, \aname achieves comparable performance to the baseline with fewer activated parameters under the same total parameter count. As the model scale increases, \aname gains a larger performance advantage over the baseline with the same total parameter count. \aname effectively leverages the increased hidden dimension redundancy in larger models, ultimately achieving higher parameter efficiency.

\begin{figure}[t]  
\centering  
\includegraphics[width=7.2cm]{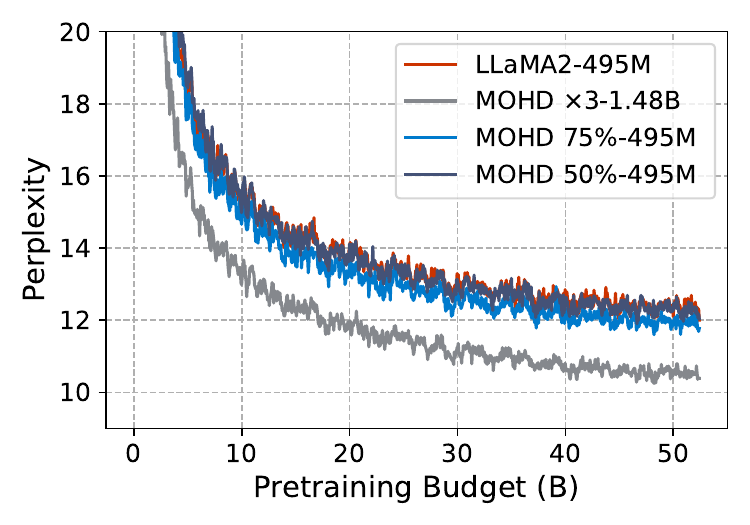}
\caption{Visualization of evaluation perplexity curves for LLaMA2-495M, \aname ×3-1.48B, \aname 75\%-495M, and \aname 50\%-495M during pretraining with 50B tokens.}
\label{fig:pt_loss}
\end{figure}

\paragraph{Training Stability.} 
In Figure~\ref{fig:pt_loss}, we visualize the evaluation perplexity curves during pretraining on 50B tokens for LLaMA2-495M, \aname ×3-1.48B, \aname 75\%-495M, and \aname 50\%-495M. The enhanced \textbf{parameter efficiency of \aname results in consistently improved training performance}, with no noticeable oscillations or anomalies. The perplexity curve for \aname ×3-1.48B, with expanded equivalent parameter counts, is lower and smoother compared to LLaMA2-495M, indicating that \aname improves the model's representation and learning capabilities. For \aname 75\%-495M and \aname 50\%-495M, the perplexity curves are slightly lower or on par with LLaMA2-495M, demonstrating that even with partial parameter activation, \aname maintains strong training characteristics. Overall, \aname effectively expands or preserves the equivalent hidden dimensions while ensuring that the representation, learning capability, and robustness during training.

\subsection{Ablation Studies}
\input{tables/ablation}

To evaluate the importance of each method in Section~\ref{sec:method} within \aname, we conducted detailed ablation experiments. In Table~\ref{tab:abla}, we compare the ablation results of \aname×2 with 710M parameters to the baseline with the same activation under zero-shot pretraining on 10B tokens, based on Eval PPL. The specific analysis is as follows:

\paragraph{Balance Loss Ablation.} The balanced loss effectively enhances \aname (0.16 improvement). It mitigates the risk of routing collapse, ensuring that most sub-dimensions are utilized more evenly. This increases the efficiency of sub-dimension utilization and improves the overall parameter efficiency of the model. For further observations and analysis on routing, see Section~\ref{sec.router.ana}.

\paragraph{Flow Maintenance Ablation.} Ablation experiments show that maintaining effective activation flow is the key factor behind \aname's high parameter efficiency. As shown in Table~\ref{tab:abla}, removing Sub-dimension Scaling leads to a significant performance drop of 1.16, indicating that without this enhancement to activation flow, the model loses a substantial amount of critical information after sparsifying the hidden dimension, making sparse \aname perform almost identically to a dense model with the same activation. Building on Sub-dimension Scaling, the Group Fusion Layer further provides a 0.22 performance gain without adding significant parameters. The Group Fusion Layer performs grouped filling and mapping after sparse activation, preserving information integrity and improving dimension utilization.

\paragraph{Mixed Activation Sub-Dimension Ablation.} In our experiments, we ablated the Mixed Activation Sub-Dimension method, using fully specialized sub-dimensions without any shared sub-dimensions. We observed a 0.83 increase in PPL, indicating a significant negative impact on model performance. This finding aligns with the observations in Section~\ref{obs.con_activate}: as there are a few common activation dimensions within the hidden layer, these dimensions should be commonly activated using a mixed activation mode, followed by sparse activation across multiple sub-dimensions. In Figure~\ref{fig:dense_total}, we also present the model's performance under various allocations of shared and specialized sub-dimensions. The mixed activation mode achieved significant performance gains over fully sparse activation, suggesting that this architecture is well-suited to the activation patterns of the Transformer model's hidden dimensions.

\subsection{Decoupled \aname Components Setting}
\input{tables/component}
\begin{figure}[t]  
\centering  
\includegraphics[width=6.5cm]{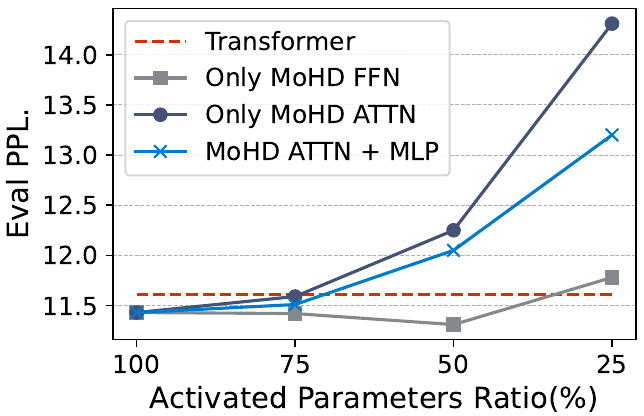}
\caption{The model's Eval PPL under different sparsity settings applied to Attention and FFN components at varying ratios.}
\label{fig:sparsification}
\vspace{-2mm}
\end{figure}
\begin{figure*}[t]
\begin{minipage}[c]{\textwidth}
\centering
  \includegraphics[width=0.98\textwidth]{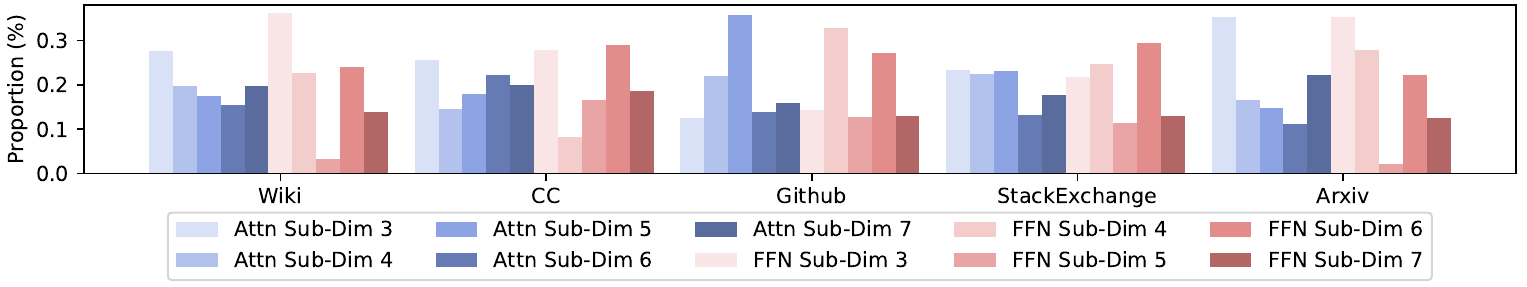}
  \caption{An illustration of Sparse Sub-dimension routing probability in \aname attention layer 1. We tested it on five domains. The bars in different colors represent the probability of different sub-dimensions being selected.}\label{fig:router_p}
\end{minipage}
\end{figure*}

To investigate the effects of sparsifying different components with \aname, we built two hidden dimension compression models based on LLaMA2-355M: one with \aname applied only to Attention and the other with \aname applied only to the FFN. Both models were trained from scratch on 10B tokens for comparison. Activation represents the model’s activation parameters, excluding the input/output embeddings. In this experiment, however, the total parameter count remains consistent across all models. To explore the impact of \aname sparsification on different components, we constructed three hidden dimension compression models based on LLaMA2-355M: one applying \aname only to Attention, only to the FFN and to both components, as shown in Figure~\ref{fig:sparsification}. Both models were pre-trained from scratch on 10B tokens for comparison. \textbf{\# Activate }here represents the model’s activation parameters, excluding input/output embeddings. The total parameter count is kept consistent across all models. The Table~\ref{tab:component} shows the effects of decoupling \aname components under three hidden dimension sparsity settings: 100\%, 50\%, and 25\%.

\paragraph{\aname Architecture Advantages.} Even with 100\% sparsity (where activation parameters match those of the original model), \aname outperformed the baseline. This may be due to \aname’s allocation of weighted activations and grouped fusion across each hidden sub-dimension, which encourages more optimal activation dimensions while suppressing noise from redundant activations, demonstrating the advantages of the \aname architecture. 

\paragraph{FFN exhibits Greater Redundancy.} The FFN layer exhibits greater redundancy in hidden dimensions, resulting in minimal performance loss (and sometimes even improvement) when sparsified. In contrast, sparsifying hidden dimensions in Attention leads to a more significant performance drop. In terms of activation parameters, the 50\% sparsity setting for the FFN uses only 195M parameters, considerably fewer than the 239M required by Attention sparsification. This suggests that the FFN is better suited for \aname transformation. From a performance perspective, the FFN achieved a PPL reduction of -0.30 in the 50\% sparsity setting, potentially due to a reduction in redundant activations that mitigates model overfitting during training, whereas Attention sparsification led to a +1.04 increase in PPL.As sparsity levels increase, the performance loss in both FFN and Attention also grows. 

\paragraph{\aname ATTN and FFN Together is Better.} Joint sparsification of Attention and FFN yields the best parameter efficiency. Under the 50\%ATTN-50\%FFN setting, the model achieved a PPL of 12.05 with only 145M activation parameters—between the 50\%FFN and 50\%ATTN configurations. Compared to applying greater sparsity to FFN alone, the 50\%ATTN-50\%FFN setting resulted in a 0.19 lower PPL than 25\%FFN, even with fewer activation parameters. This may be because consistency in activated hidden dimensions helps the model maintain better learning capacity.

\subsection{Analysis}\label{sec.router.ana}

\begin{figure}[t]
\centering
\begin{subfigure}{0.48\linewidth}
    \centering
    \includegraphics[width=\linewidth]{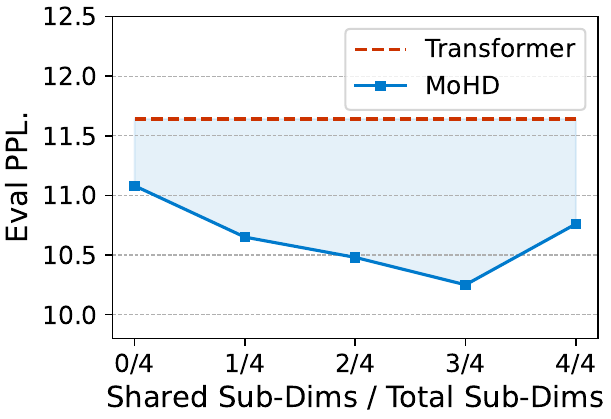}
    \caption{Eval PPL. under different Shared Sub-Dim ratio settings.}
    \label{fig:dense_total}
\end{subfigure}
\hspace{0.02cm}
\begin{subfigure}{0.48\linewidth}
    \centering
    \includegraphics[width=\linewidth]{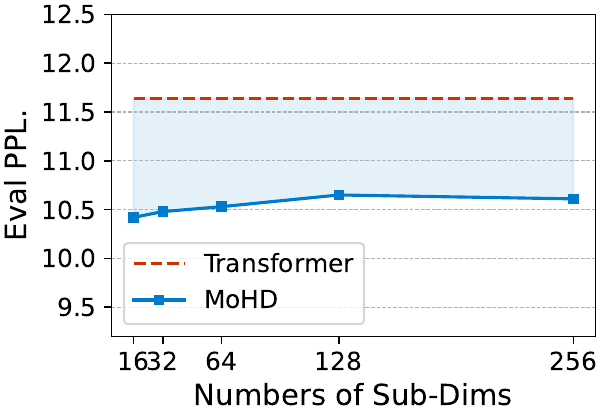}
    \caption{Eval PPL. under different Sub-Dim Numbers settings.}
    \label{fig:nums_subdim}
\end{subfigure}
\caption{Performance of \aname×2-710M with varying sub-dimension allocation ratios and finer-grained sub-dimension settings. All models are pre-trained from scratch on 10B tokens.}
\label{ablation study figures}
\end{figure}

\paragraph{Router Probability.} To observe the targeted sub-dimension selection based on the router in \aname, we visualize the attention and FFN router weight distributions at the 5th layer of \aname across five different data domains in Figure~\ref{fig:router_p}. Each probability weight represents the average selection probability across 4096 tokens. We can observe that sub-dimensions show specialization across different data domains. For instance, in the Attention \aname Router, Subdim 5 demonstrates the importance of code-related data, with significantly higher probabilities in the GitHub and StackExchange domains. On the other hand, Attention Subdim 3 shows higher probabilities in Wiki, CC, and ArXiv domains, while it is much lower in GitHub. We speculate that this Subdim is important for commonsense knowledge and writing tasks. In the FFN \aname Router, Subdim 4 specializes in code-related tasks, while Subdim 3 specializes in commonsense knowledge tasks. The specialization of sub-dimensions validates the effectiveness of \aname, enabling it to allocate differentiated sub-dimensions to individual tokens based on various data entries and domains. As a result, \aname can effectively leverage these sub-dimensions to increase the equivalent parameter count, achieving higher parameter efficiency. Furthermore, while the probabilities of all sub-dimensions in \aname differ, they generally stay within the 0.2-0.3 range, indicating that all sub-dimensions are actively chosen, thus preventing router collapse.

\paragraph{Shared Activation v.s. Specialized Activation.} Figure~\ref{fig:dense_total} shows the evaluation PPL values of \aname×2-710M trained on 10B data under different Shared activation and mid-activation proportions. The non-activation parameters of the Baseline and \aname models are identical. Firstly, \aname routing effectively increases the model's equivalent hidden dimensions. In the 0/4 setting, the routing yields better performance compared to the baseline, while in the 4/4 setting, it results in a significant performance drop. However, retaining Shared activation in the hidden dimensions proves essential. We found that the best performance occurs when Shared activation is set to 3/4, which is why this ratio was commonly used in the experiments. This indicates that the routing design for hidden dimensions still requires further research, and there is ample room for exploration to increase the model's sparsity.

\paragraph{Fine Gain Sub-dim Dimension.} Figure~\ref{fig:nums_subdim} presents the test PPL values of \aname×2 -710M after pre-training on 10B data, with a finer-grained division of sub-dimensions. When the number of sub-dimensions is set to 16 (sub-dimension size 256), the model achieves the best performance. As the number of sub-dimensions increases to 128 (sub-dimension size 32), the model’s performance slightly improves, and a further increase to 256 sub-dimensions (sub-dimension size 16) results in a marginal improvement. This experiment demonstrates that, within the current \aname design, increasing the number of sub-dimensions does not enhance model performance but instead leads to higher computational costs, especially in the routing and grouping fusion layers. The results also validate the effectiveness of the grouping fusion layer: fine-grained grouping fusion mechanisms are not necessary, and a small number of parameters are sufficient to maintain effective forward activation flow.

\section{Related Work}

\subsection{Activation Sparsity} Activation sparsity refers to the phenomenon where a significant proportion of a model's hidden states are zero-valued. This property naturally arises in the intermediate states of ReLU-based MLPs, as demonstrated in prior work \cite{you2022lazyneuron,li2023lazy}. Some studies have leveraged activation sparsity to improve the efficiency of LLMs during inference. \citet{liu2023deja} utilized activation sparsity to accelerate LLM inference by omitting the transfer of weight channels corresponding to zero-valued entries to GPU registers. Additionally, \citet{song2023powerinfer} and \citet{alizadeh2024llmflash} extended this concept to CPU offloading, significantly reducing memory transfer overhead between CPUs and GPUs. Recent works has reintroduced activation sparsity into LLM architectures to enhance efficiency. \citet{mirzadeh2023relu} replaced SiLU and GeLU with ReLU, achieving sparsity through extended pretraining. \citet{zhang2024relu2} identified Squared ReLU \cite{so2022squaredrelu} as a superior alternative for sparse activations.  \citet{song2024prosparse, song2024activation} proposed regularization techniques to increase sparsity, while  \citet{wang2024ladder} combined pruning and quantized activations to establish scaling laws. \citet{lee2024cats} introduced CATS, achieving training-free sparsity in SwiGLU-based LLMs. \citet{liu2024trainingfreeactivationsparsitylarge} extended these concepts to training-free activation sparsity for large-scale language models. Building on prior studies, we investigate hidden dimension sparsity, focusing on continuous activation across tokens. Leveraging this, we design a sparse activation architecture that improves parameter efficiency and enhances hidden dimension scalability.

\subsection{Sparsely-activated Transformer}
Sparsely-activated Transformer models, such as Sparse Mixture-of-Expert (MoE) architectures, leverage input adaptivity to achieve scalable and efficient computation. These models dynamically activate only a subset of specialized subnetworks, or "experts," for processing each input token, significantly reducing computational overhead \cite{fedus2022switch, riquelme2021scaling, zhou2022mixture, jiang2024mixtral,xue2024openmoe}. This mechanism enables effective handling of diverse data domains \cite{li2022branch,jain2024nested} while maintaining high performance. Recent advancements in sparsely-activated Transformers have extended their capabilities by introducing heterogeneous experts~\cite{wu2024multihead,he2024millionexperts}, allowing networks to integrate experts with varying capacities and specializations \cite{ dean2021pathways, zhou2022mixture,dai2024deepseekmoeultimateexpertspecialization}. Some recent studies\cite{qiu-etal-2024-unlocking} have observed the activation patterns in the intermediate dimensions of FFNs and explored sparsely-activated architectures based on these observations. However, no existing Transformer architecture has implemented sparse activation specifically in the hidden dimensions. Inspired by the work of~\cite{qiu-etal-2024-unlocking,dai2024deepseekmoeultimateexpertspecialization}, we conducted an in-depth analysis of the hidden dimensions and designed a novel sparse activation strategy. This innovation opens a new research avenue for sparsely-activated Transformer architectures.


\section{Conclusion}

In this paper, we presented \aname(Mixture of Hidden Dimensions), a sparse conditional activation architecture designed to address the inefficiencies in scaling Transformer hidden dimensions. By integrating shared sub-dimensions for common token features and dynamically activating specialized sub-dimensions through a routing mechanism, MOHD achieves improved efficiency and flexibility while preserving activation flow through activation scaling and group fusion mechanisms. Our evaluations demonstrate that MOHD outperforms standard Transformers across multiple NLP tasks, achieving superior parameter efficiency and enhanced task performance. These results underscore the potential of hidden dimension sparsity as a promising direction for improving the scalability and efficiency of Transformer.

\section*{Acknowledgments}
We would like to thank Yinqi Yang, Yanxi Xie, Naibin Gu, Kun Huang and members of the IIE KDsec NLP group for their valuable feedback and discussions. We are very grateful to Mengzhou Xia for providing the concise and effective ShearingLLaMA experimental code and for her assistance during the reproduction process. Work done during Yilong Chen’s internship in Baidu Inc. 
\nocite{langley00}

\bibliography{icml_2025}
\bibliographystyle{icml2024}

\newpage



\end{document}

%% file: tables/model_setting.tex
\begin{table}[t]
\centering
\caption{Detailed configuration, activation parameters, and total parameters of the models included in our study. L.2-355M represents the LaMMA-2 architecture model with 355M total parameters.}
\vskip 0.15in
\small{\resizebox{0.9\columnwidth}{!}{%
\begin{tabular}{@{}lcccc@{}}
\toprule
\textbf{Model Setting} & \textbf{L.2-355M} &  \textbf{L.2-495M} &   \textbf{L.2-1.13B} \\ \midrule
\textit{hidden size }        & 1024  & 1536 & 2048  \\
\textit{intermediate size }        &  2560  & 2560 & 4096  \\
\textit{attention heads}         &  32  & 32 & 32  \\
\textit{num kv heads}      &  32  & 16 & 32 \\
\textit{layers }         & 24  & 24 & 24\\
\midrule
\textbf{\# Activate}  & 289M   & 396M & 1B \\
\textbf{\# Params}  & 355M   & 495M & 1.13B \\
\bottomrule
\end{tabular}
}}

\label{tab:methods}
\end{table}

%% file: tables/main.tex
\begin{table*}[!ht]
\caption{Comprehensively evaluate the basic capabilities of models with different activation parameters. In particular, \aname 50\%-355M represents a model with 355M total parameters using \aname to compress 50\% hidden dimensions. 
Abbreviated task names LM stands for language modeling, WG stands for Winogrande, Hella stands for Hellaswag, and Lam stands for Lambada. Green and red values indicate metrics that exceed or fall below the baseline, respectively. \textbf{\# Activate} refers to all activation parameters excluding the Embedding layers.}
\vskip 0.15in
\setlength\tabcolsep{5.5pt}
\footnotesize
    \centering
    
    \begin{tabular*}{1.0\textwidth}{@{\extracolsep{\fill}}@{}l c cccccc cc cc c @{}}
    \toprule
    \multirow{2}{*}{\raisebox{-0.5\height}{\textbf{Model-Params}}} & \multirow{2}{*}{\raisebox{-0.5\height}{\textbf{\# Activate}}} & \multicolumn{6}{c}{\scriptsize\textbf{Commonsense \& Reading Comprehension}} & \multicolumn{2}{c}{\scriptsize\textbf{Continued}} &  \multicolumn{1}{c}{\scriptsize\textbf{LM}} & \multicolumn{1}{c}{\scriptsize\textbf{Knowledge}}  & \multirow{2}{*}{\raisebox{-0.5\height}{\textbf{Avg.}}} \\
    \cmidrule(lr){3-8} \cmidrule(lr){9-10} \cmidrule(lr){11-11} \cmidrule(lr){12-12}
    & & \scriptsize\textbf{SciQ} & \scriptsize\textbf{PIQA} & \scriptsize\textbf{WG} & \scriptsize\textbf{ARC-E } & \scriptsize\textbf{ARC-C} & \scriptsize\textbf{Hella.} & \scriptsize\textbf{LogiQA}  & \scriptsize\textbf{BoolQ} & \scriptsize\textbf{Lam.} & \scriptsize\textbf{MMLU} \\
    \midrule
    LLaMA2-355M & 289M & 74.0 & 65.2 & 50.5 & 44.7 & 20.1 & 31.1 & 19.5 & 59.7 & 36.6 & 25.2 & 42.7 \\
    \midrule
     \rowcolor{myblue!50} \aname$~50\%$-355M & 145M & 74.0 & \textcolor{green!60!black}{65.6} & \textcolor{red!60!black}{50.4} & \textcolor{red!60!black}{43.9} & \textcolor{red!60!black}{19.7} & \textcolor{red!60!black}{30.7} & \textcolor{green!60!black}{20.6} & \textcolor{red!60!black}{54.7} & \textcolor{green!60!black}{37.7} & \textcolor{green!60!black}{25.6} & \textcolor{red!60!black}{42.3} \\
     \rowcolor{myblue!50}   \aname$~75\%$-355M & 217M & \textcolor{green!60!black}{75.3} & \textcolor{green!60!black}{65.6} & \textcolor{green!60!black}{50.9} & \textcolor{green!60!black}{44.7} & \textcolor{green!60!black}{20.9} & \textcolor{green!60!black}{31.1} & \textcolor{green!60!black}{22.3} & \textcolor{red!60!black}{55.8} & \textcolor{green!60!black}{38.9} & \textcolor{green!60!black}{26.2} & \textcolor{green!60!black}{43.2} \\
    \rowcolor{myred!15}    \aname$\times2$ -710M & 289M & \textcolor{green!60!black}{76.6} & \textcolor{green!60!black}{67.5} & \textcolor{red!60!black}{49.8} & \textcolor{green!60!black}{47.7} & \textcolor{green!60!black}{23.0} & \textcolor{green!60!black}{33.4} & \textcolor{green!60!black}{20.7} & \textcolor{green!60!black}{60.5} & \textcolor{green!60!black}{43.3} & \textcolor{green!60!black}{26.5} & \textcolor{green!60!black}{44.9} \\
    \rowcolor{myred!15}    \aname$\times3$ -1.06B & 289M & \textcolor{green!60!black}{77.1} & \textcolor{green!60!black}{67.8} & \textcolor{green!60!black}{51.1} & \textcolor{green!60!black}{47.6} & \textcolor{green!60!black}{21.8} & \textcolor{green!60!black}{33.9} & \textcolor{green!60!black}{20.6} & \textcolor{red!60!black}{55.8} & \textcolor{green!60!black}{43.6} & \textcolor{green!60!black}{25.6} & \textcolor{green!60!black}{44.5} \\
    \rowcolor{myred!15}    \aname$\times4$ -1.42B & 289M & \textcolor{green!60!black}{77.6} & \textcolor{green!60!black}{67.9} & \textcolor{red!60!black}{49.1} & \textcolor{green!60!black}{47.0} & \textcolor{green!60!black}{23.3} & \textcolor{green!60!black}{33.9} & \textcolor{green!60!black}{22.1} & \textcolor{red!60!black}{57.5} & \textcolor{green!60!black}{44.3} & \textcolor{red!60!black}{24.7} & \textcolor{green!60!black}{44.7} \\

    \midrule
    LLaMA2-495M & 396M &  75.4 & 66.5 & 51.3 & 45.5 & 19.9 & 32.0 & 21.7 & 60.5 & 38.9 & 25.8 & 43.8 \\
    \midrule
     \rowcolor{myblue!50} 
        \aname$~50\%$-495M & 198M & \textcolor{green!60!black}{76.9} & \textcolor{green!60!black}{67.1} & \textcolor{green!60!black}{52.7} & \textcolor{green!60!black}{46.4} & \textcolor{green!60!black}{20.1} & \textcolor{green!60!black}{32.3} & \textcolor{red!60!black}{21.5} & \textcolor{red!60!black}{57.0} & \textcolor{green!60!black}{40.7} & \textcolor{green!60!black}{26.2} & \textcolor{green!60!black}{44.1} \\
        \rowcolor{myblue!50} 
        \aname$~75\%$-495M & 297M & \textcolor{green!60!black}{76.4} & \textcolor{green!60!black}{67.3} & \textcolor{red!60!black}{50.6} & \textcolor{green!60!black}{45.8} & \textcolor{green!60!black}{21.1} & \textcolor{green!60!black}{33.0} & \textcolor{green!60!black}{23.7} & \textcolor{green!60!black}{61.8} & \textcolor{green!60!black}{41.7} & \textcolor{green!60!black}{26.2} & \textcolor{green!60!black}{44.8} \\

        \rowcolor{myred!15} 
        \aname$\times2$ -989M & 396M & \textcolor{green!60!black}{77.1} & \textcolor{green!60!black}{67.8} & \textcolor{red!60!black}{51.1} & \textcolor{green!60!black}{47.6} & \textcolor{green!60!black}{21.8} & \textcolor{green!60!black}{33.9} & \textcolor{red!60!black}{20.6} & \textcolor{red!60!black}{55.8} & \textcolor{green!60!black}{43.6} & \textcolor{red!60!black}{25.6} & \textcolor{green!60!black}{44.5} \\
        \rowcolor{myred!15} 
        \aname$\times3$ -1.48B & 396M & \textcolor{green!60!black}{77.0} & \textcolor{green!60!black}{69.0} & \textcolor{red!60!black}{51.1} & \textcolor{green!60!black}{48.8} & \textcolor{green!60!black}{23.6} & \textcolor{green!60!black}{35.6} & \textcolor{green!60!black}{22.0} & \textcolor{red!60!black}{58.6} & \textcolor{green!60!black}{48.2} & \textcolor{green!60!black}{26.1} & \textcolor{green!60!black}{46.0} \\
         \rowcolor{myred!15} \aname$\times4$ -1.98B & 396M & \textcolor{green!60!black}{79.1} & \textcolor{green!60!black}{67.4} & \textcolor{red!60!black}{49.8} & \textcolor{green!60!black}{49.1} & \textcolor{green!60!black}{22.0} & \textcolor{green!60!black}{35.2} & \textcolor{red!60!black}{20.7} & \textcolor{green!60!black}{60.9} & \textcolor{green!60!black}{47.4} & \textcolor{green!60!black}{26.1} & \textcolor{green!60!black}{45.8} \\
    \midrule
    LLaMA2-1.13B & 1B & 81.0 & 68.1 & 51.8 & 49.3 & 23.2 & 35.0 & 21.7 & 47.0 & 38.9 & 26.4 & 44.2 \\
    \midrule
        \rowcolor{myblue!50} 
        \aname$~50\%$-1.13B & 503M & \textcolor{red!60!black}{78.9} & \textcolor{red!60!black}{67.8} & \textcolor{red!60!black}{50.1} & \textcolor{red!60!black}{48.7} & \textcolor{red!60!black}{21.2} & \textcolor{green!60!black}{35.2} & \textcolor{red!60!black}{21.5} & \textcolor{green!60!black}{61.1} & \textcolor{green!60!black}{48.8} & \textcolor{red!60!black}{25.5} & \textcolor{green!60!black}{45.9} \\
        \rowcolor{myblue!50} 
        \aname$~75\%$-1.13B & 755M & \textcolor{green!60!black}{80.3} & \textcolor{green!60!black}{69.3} & \textcolor{green!60!black}{52.3} & \textcolor{green!60!black}{50.8} & \textcolor{green!60!black}{24.5} & \textcolor{green!60!black}{36.1} & \textcolor{green!60!black}{22.3} & \textcolor{green!60!black}{51.2} & \textcolor{green!60!black}{48.4} & \textcolor{red!60!black}{25.0} & \textcolor{green!60!black}{46.0} \\
        \rowcolor{myred!15} 
        \aname$\times2$ -2.27B & 1B & \textcolor{green!60!black}{81.2} & \textcolor{green!60!black}{70.9} & \textcolor{green!60!black}{54.1} & \textcolor{green!60!black}{53.0} & \textcolor{green!60!black}{24.6} & \textcolor{green!60!black}{38.3} & \textcolor{green!60!black}{22.4} & \textcolor{green!60!black}{50.5} & \textcolor{green!60!black}{52.1} & \textcolor{red!60!black}{25.5} & \textcolor{green!60!black}{47.2} \\
        \rowcolor{myred!15} 
        \aname$\times3$ -3.41B & 1B & \textcolor{green!60!black}{83.6} & \textcolor{green!60!black}{69.8} & \textcolor{green!60!black}{53.1} & \textcolor{green!60!black}{51.9} & \textcolor{green!60!black}{25.4} & \textcolor{green!60!black}{38.3} & \textcolor{red!60!black}{21.0} & \textcolor{green!60!black}{56.5} & \textcolor{green!60!black}{53.0} & \textcolor{green!60!black}{26.6} & \textcolor{green!60!black}{47.9} \\
        \rowcolor{myred!15} \aname$\times4$ -4.55B & 1B & \textcolor{green!60!black}{82.4} & \textcolor{green!60!black}{70.0} & \textcolor{green!60!black}{52.8} & \textcolor{green!60!black}{51.6} & \textcolor{green!60!black}{23.4} & \textcolor{green!60!black}{38.0} & \textcolor{green!60!black}{23.4} & \textcolor{green!60!black}{54.9} & \textcolor{green!60!black}{50.7} & \textcolor{green!60!black}{26.6} & \textcolor{green!60!black}{47.4} \\
     \midrule
    LLaMA2-2.7b & 2.54B & 82.5 & 70.8 & 56.3 & 54.4 & 27.8 & 39.3 & 23.5 & 44.4 & 37.7 & 25.3 & 46.2 \\
    \bottomrule
    \end{tabular*}


    \label{table:main-results1}

\vspace{-0.4cm}
\end{table*}

%% file: tables/moh_setting.tex
\begin{table}[t]
\centering
\caption{Parameter configurations of MoHD under compression and expansion experiments. We use the same settings for both Attention and FFN. For detailed reasoning behind these configurations, please refer to Analysis.}
\vskip 0.15in
\small{\resizebox{0.9\columnwidth}{!}{%
\begin{tabular}{@{}lcccccc@{}}
\toprule
\textbf{MoH} & \textbf{$~50\%$} &  \textbf{$~75\%$} &   \textbf{$\times2$} &   \textbf{$\times3$} &   \textbf{$\times4$} \\ \midrule
\textit{attn top k }        & 4  & 4 & 4 & 4 & 4  \\
\textit{attn sub-dim num}        &  8  & 12 & 8 & 12 & 16\\
\textit{ffn top k}          & 4  & 4 & 4 & 4 & 4  \\
\textit{ffn sub-dim num}     &  8  & 12 & 8 & 12 & 16  \\
\textit{shared sub-dim num}         & 3  & 3 & 3 & 3 & 3\\
\textit{group fusion dim}         & 8  & 12 & 8 & 12 & 16\\
\bottomrule
\end{tabular}
}}

\label{tab:methods}
\end{table}

%% file: tables/ablation.tex
\begin{table}[t]
\centering
\caption{Eval Perplexity with ablation on 10B training MoH. "w.o." indicates the method was ablated based on MoH×2-710M.}
\vskip 0.15in
\small
\setlength{\tabcolsep}{4.5mm}{
\begin{tabular}{@{}lc@{}}
\toprule
\textbf{Method} & \textbf{Perplexity} $\downarrow$  \\
 \midrule
 MoH×2 -710M & 10.25 \\
   \midrule
w.o. Mixed Activated Sub-Dimensions & 11.08 (\textcolor{red!50!black}{+0.83}) \\
w.o. Balance Loss & 10.41 (\textcolor{red!50!black}{+0.16}) \\
w.o. Group Fusion Layer & 10.47 (\textcolor{red!50!black}{+0.22}) \\
w.o. Sub-Dimension Scaling & 11.41 (\textcolor{red!50!black}{+1.16}) \\
\midrule
LLaMA2-355M & 11.61\textcolor{red!50!black}{(+1.36)} \\
 \bottomrule
\end{tabular}
}
\label{tab:abla}
\end{table}

%% file: tables/component.tex
\begin{table}[t]
\centering
\caption{Eval Perplexity in the MoH setting is performed for the Attention or FFN of LLaMA2-355M. All models were pre-trained on 10B data after initialization. \# Activation represents the activation parameter of the model, excluding the input/output Embedding.}
\vskip 0.15in
\small
\setlength{\tabcolsep}{3mm}{
\begin{tabular}{@{}lcc@{}}
\toprule
\textbf{Method} & \textbf{\# Activate} & \textbf{Perplexity} $\downarrow$\\
 \midrule
 LLaMA2-355M& 289M &11.61\\
\midrule
MoH-100\%ATTN-100\%FFN& 289M         &11.43 (\textcolor{green!70!black}{-0.18})\\
\midrule
MoH-100\%ATTN-50\%FFN& 195M         &11.31 (\textcolor{green!70!black}{-0.30})\\
MoH-50\%ATTN-100\%FFN& 239M       & 12.25 (\textcolor{red!70!black}{+0.64})\\
MoH-50\%ATTN-50\%FFN& 145M          & 12.05 (\textcolor{red!70!black}{+0.44})\\
\midrule
 MoH-100\%ATTN-25\%FFN& 147M&12.24 (\textcolor{red!70!black}{+0.63})\\
MoH-25\%ATTN-100\%FFN& 213M&14.31 (\textcolor{red!70!black}{+2.70})\\
MoH-25\%ATTN-25\%FFN& 72M          & 13.20 (\textcolor{red!70!black}{+1.59})\\
\bottomrule
\end{tabular}
}
\label{tab:component}
\end{table}

%% file: icml_2025.bbl
\begin{thebibliography}{53}
\providecommand{\natexlab}[1]{#1}
\providecommand{\url}[1]{\texttt{#1}}
\expandafter\ifx\csname urlstyle\endcsname\relax
  \providecommand{\doi}[1]{doi: #1}\else
  \providecommand{\doi}{doi: \begingroup \urlstyle{rm}\Url}\fi

\bibitem[Alizadeh et~al.(2024)Alizadeh, Mirzadeh, Belenko, Khatamifard, Cho, Del~Mundo, Rastegari, and Farajtabar]{alizadeh2024llmflash}
Alizadeh, K., Mirzadeh, I., Belenko, D., Khatamifard, K., Cho, M., Del~Mundo, C.~C., Rastegari, M., and Farajtabar, M.
\newblock Llm in a flash: Efficient large language model inference with limited memory.
\newblock \emph{arXiv preprint arXiv:2312.11514}, 2024.
\newblock URL \url{https://arxiv.org/abs/2312.11514}.

\bibitem[Anthropic(2023)]{claude}
Anthropic.
\newblock Anthropic: Introducing claude 2.1, 2023.
\newblock URL \url{https://www.anthropic.com/index/claude-2-1}.

\bibitem[Bisk et~al.(2020)Bisk, Zellers, Gao, Choi, et~al.]{piqa}
Bisk, Y., Zellers, R., Gao, J., Choi, Y., et~al.
\newblock Piqa: Reasoning about physical commonsense in natural language.
\newblock In \emph{Proceedings of the AAAI conference on artificial intelligence}, volume~34, pp.\  7432--7439, 2020.

\bibitem[Cai et~al.(2024{\natexlab{a}})Cai, Muralidharan, Heinrich, Yin, Wang, Kautz, and Molchanov]{cai2024flextronmanyinoneflexiblelarge}
Cai, R., Muralidharan, S., Heinrich, G., Yin, H., Wang, Z., Kautz, J., and Molchanov, P.
\newblock Flextron: Many-in-one flexible large language model, 2024{\natexlab{a}}.
\newblock URL \url{https://arxiv.org/abs/2406.10260}.

\bibitem[Cai et~al.(2024{\natexlab{b}})Cai, Jiang, Wang, Tang, Kim, and Huang]{cai2024surveymixtureexperts}
Cai, W., Jiang, J., Wang, F., Tang, J., Kim, S., and Huang, J.
\newblock A survey on mixture of experts, 2024{\natexlab{b}}.
\newblock URL \url{https://arxiv.org/abs/2407.06204}.

\bibitem[Chen et~al.(2023)Chen, Ding, Yadav, Zharkov, and Liang]{chenLoRAShearEfficientLarge2023}
Chen, T., Ding, T., Yadav, B., Zharkov, I., and Liang, L.
\newblock {LoRAShear}: {Efficient} {Large} {Language} {Model} {Structured} {Pruning} and {Knowledge} {Recovery}, October 2023.
\newblock URL \url{https://arxiv.org/abs/2310.18356v2}.

\bibitem[Chen et~al.(2024)Chen, Shang, Zhang, Cui, Liu, Wang, Sun, and Wu]{chen-etal-2024-lemon}
Chen, Y., Shang, J., Zhang, Z., Cui, S., Liu, T., Wang, S., Sun, Y., and Wu, H.
\newblock {LEMON}: Reviving stronger and smaller {LM}s from larger {LM}s with linear parameter fusion.
\newblock In Ku, L.-W., Martins, A., and Srikumar, V. (eds.), \emph{Proceedings of the 62nd Annual Meeting of the Association for Computational Linguistics (Volume 1: Long Papers)}, pp.\  8005--8019, Bangkok, Thailand, August 2024. Association for Computational Linguistics.
\newblock \doi{10.18653/v1/2024.acl-long.434}.
\newblock URL \url{https://aclanthology.org/2024.acl-long.434}.

\bibitem[Clark et~al.(2019)Clark, Lee, Chang, Kwiatkowski, Collins, and Toutanova]{clark2019boolq}
Clark, C., Lee, K., Chang, M.-W., Kwiatkowski, T., Collins, M., and Toutanova, K.
\newblock Boolq: Exploring the surprising difficulty of natural yes/no questions.
\newblock \emph{arXiv preprint arXiv:1905.10044}, 2019.

\bibitem[Clark et~al.(2018{\natexlab{a}})Clark, Cowhey, Etzioni, Khot, Sabharwal, Schoenick, and Tafjord]{arcChallenge:journals/corr/abs-1803-05457}
Clark, P., Cowhey, I., Etzioni, O., Khot, T., Sabharwal, A., Schoenick, C., and Tafjord, O.
\newblock Think you have solved question answering? try arc, the {AI2} reasoning challenge.
\newblock \emph{CoRR}, abs/1803.05457, 2018{\natexlab{a}}.
\newblock URL \url{http://arxiv.org/abs/1803.05457}.

\bibitem[Clark et~al.(2018{\natexlab{b}})Clark, Cowhey, Etzioni, Khot, Sabharwal, Schoenick, and Tafjord]{clark2018think}
Clark, P., Cowhey, I., Etzioni, O., Khot, T., Sabharwal, A., Schoenick, C., and Tafjord, O.
\newblock Think you have solved question answering? try arc, the ai2 reasoning challenge.
\newblock \emph{arXiv preprint arXiv:1803.05457}, 2018{\natexlab{b}}.

\bibitem[Dai et~al.(2024)Dai, Deng, Zhao, Xu, Gao, Chen, Li, Zeng, Yu, Wu, Xie, Li, Huang, Luo, Ruan, Sui, and Liang]{dai2024deepseekmoeultimateexpertspecialization}
Dai, D., Deng, C., Zhao, C., Xu, R.~X., Gao, H., Chen, D., Li, J., Zeng, W., Yu, X., Wu, Y., Xie, Z., Li, Y.~K., Huang, P., Luo, F., Ruan, C., Sui, Z., and Liang, W.
\newblock Deepseekmoe: Towards ultimate expert specialization in mixture-of-experts language models, 2024.
\newblock URL \url{https://arxiv.org/abs/2401.06066}.

\bibitem[Dao et~al.(2022)Dao, Chen, Sohoni, Desai, Poli, Grogan, Liu, Rao, Rudra, and Ré]{daoMonarchExpressiveStructured2022a}
Dao, T., Chen, B., Sohoni, N., Desai, A., Poli, M., Grogan, J., Liu, A., Rao, A., Rudra, A., and Ré, C.
\newblock Monarch: {Expressive} {Structured} {Matrices} for {Efficient} and {Accurate} {Training}, April 2022.
\newblock URL \url{http://arxiv.org/abs/2204.00595}.
\newblock arXiv:2204.00595 [cs].

\bibitem[Dean(2021)]{dean2021pathways}
Dean, J.
\newblock Introducing pathways: A next-generation ai architecture.
\newblock \emph{Google Blog}, 366, 2021.

\bibitem[Fedus et~al.(2022)Fedus, Zoph, and Shazeer]{fedus2022switch}
Fedus, W., Zoph, B., and Shazeer, N.
\newblock Switch transformers: Scaling to trillion parameter models with simple and efficient sparsity.
\newblock \emph{The Journal of Machine Learning Research}, 23\penalty0 (1):\penalty0 5232--5270, 2022.

\bibitem[Gao et~al.(2021)Gao, Tow, Biderman, Black, DiPofi, Foster, Golding, Hsu, McDonell, Muennighoff, Phang, Reynolds, Tang, Thite, Wang, Wang, and Zou]{eval-harness}
Gao, L., Tow, J., Biderman, S., Black, S., DiPofi, A., Foster, C., Golding, L., Hsu, J., McDonell, K., Muennighoff, N., Phang, J., Reynolds, L., Tang, E., Thite, A., Wang, B., Wang, K., and Zou, A.
\newblock A framework for few-shot language model evaluation.
\newblock In \emph{Zenodo}. https://doi.org/10.5281/zenodo.5371628, September 2021.

\bibitem[He(2024)]{he2024millionexperts}
He, X.
\newblock Mixture of a million experts.
\newblock \emph{arXiv preprint arXiv:2407.04153}, 2024.
\newblock URL \url{https://arxiv.org/abs/2407.04153}.

\bibitem[Jain et~al.(2024)Jain, Hegde, Kusupati, Nagrani, and Buch]{jain2024nested}
Jain, G., Hegde, N., Kusupati, A., Nagrani, A., and Buch, S.
\newblock Mixture of nested experts: Adaptive processing of visual tokens.
\newblock \emph{arXiv preprint arXiv:2407.19985}, 2024.
\newblock URL \url{https://arxiv.org/abs/2407.19985}.

\bibitem[Jiang et~al.(2024{\natexlab{a}})Jiang, Sablayrolles, Roux, Mensch, Savary, Bamford, Chaplot, Casas, Hanna, Bressand, et~al.]{jiang2024mixtral}
Jiang, A.~Q., Sablayrolles, A., Roux, A., Mensch, A., Savary, B., Bamford, C., Chaplot, D.~S., Casas, D. d.~l., Hanna, E.~B., Bressand, F., et~al.
\newblock Mixtral of experts.
\newblock \emph{arXiv preprint arXiv:2401.04088}, 2024{\natexlab{a}}.

\bibitem[Jiang et~al.(2024{\natexlab{b}})Jiang, Sablayrolles, Roux, Mensch, Savary, Bamford, Chaplot, de~las Casas, Hanna, Bressand, Lengyel, Bour, Lample, Lavaud, Saulnier, Lachaux, Stock, Subramanian, Yang, Antoniak, Scao, Gervet, Lavril, Wang, Lacroix, and Sayed]{jiang2024mixtralexperts}
Jiang, A.~Q., Sablayrolles, A., Roux, A., Mensch, A., Savary, B., Bamford, C., Chaplot, D.~S., de~las Casas, D., Hanna, E.~B., Bressand, F., Lengyel, G., Bour, G., Lample, G., Lavaud, L.~R., Saulnier, L., Lachaux, M.-A., Stock, P., Subramanian, S., Yang, S., Antoniak, S., Scao, T.~L., Gervet, T., Lavril, T., Wang, T., Lacroix, T., and Sayed, W.~E.
\newblock Mixtral of experts, 2024{\natexlab{b}}.
\newblock URL \url{https://arxiv.org/abs/2401.04088}.

\bibitem[Kaplan et~al.(2020)Kaplan, McCandlish, Henighan, Brown, Chess, Child, Gray, Radford, Wu, and Amodei]{kaplan2020scalinglawsneurallanguage}
Kaplan, J., McCandlish, S., Henighan, T., Brown, T.~B., Chess, B., Child, R., Gray, S., Radford, A., Wu, J., and Amodei, D.
\newblock Scaling laws for neural language models, 2020.
\newblock URL \url{https://arxiv.org/abs/2001.08361}.

\bibitem[Lee et~al.(2024)]{lee2024cats}
Lee, J. et~al.
\newblock Cats: Training-free activation sparsity for swiglu-based llms.
\newblock \emph{arXiv preprint}, 2024.
\newblock URL \url{https://arxiv.org/abs/2401.12345}.

\bibitem[Li et~al.(2022)Li, Gururangan, Dettmers, Lewis, Althoff, Smith, and Zettlemoyer]{li2022branch}
Li, M., Gururangan, S., Dettmers, T., Lewis, M., Althoff, T., Smith, N.~A., and Zettlemoyer, L.
\newblock Branch-train-merge: Embarrassingly parallel training of expert language models.
\newblock \emph{arXiv preprint arXiv:2208.03306}, 2022.

\bibitem[Li et~al.(2023)Li, You, Bhojanapalli, Li, Rawat, Reddi, Ye, Chern, Yu, Guo, and Kumar]{li2023lazy}
Li, Z., You, C., Bhojanapalli, S., Li, D., Rawat, A.~S., Reddi, S.~J., Ye, K., Chern, F., Yu, F., Guo, R., and Kumar, S.
\newblock The lazy neuron phenomenon: On emergence of activation sparsity in transformers.
\newblock \emph{arXiv preprint arXiv:2210.06313}, 2023.
\newblock URL \url{https://arxiv.org/abs/2210.06313}.

\bibitem[Liu et~al.(2020)Liu, Cui, Liu, Huang, Wang, and Zhang]{liu2020logiqa}
Liu, J., Cui, L., Liu, H., Huang, D., Wang, Y., and Zhang, Y.
\newblock Logiqa: A challenge dataset for machine reading comprehension with logical reasoning.
\newblock \emph{arXiv preprint arXiv:2007.08124}, 2020.

\bibitem[Liu et~al.(2024)Liu, Ponnusamy, Cai, Guo, Kim, and Athiwaratkun]{liu2024trainingfreeactivationsparsitylarge}
Liu, J., Ponnusamy, P., Cai, T., Guo, H., Kim, Y., and Athiwaratkun, B.
\newblock Training-free activation sparsity in large language models, 2024.
\newblock URL \url{https://arxiv.org/abs/2408.14690}.

\bibitem[Liu et~al.(2023{\natexlab{a}})Liu, Wang, Dao, Zhou, Yuan, Song, Shrivastava, Zhang, Tian, Re, and Chen]{dejiavu}
Liu, Z., Wang, J., Dao, T., Zhou, T., Yuan, B., Song, Z., Shrivastava, A., Zhang, C., Tian, Y., Re, C., and Chen, B.
\newblock Deja vu: Contextual sparsity for efficient {LLM}s at inference time.
\newblock In Krause, A., Brunskill, E., Cho, K., Engelhardt, B., Sabato, S., and Scarlett, J. (eds.), \emph{Proceedings of the 40th International Conference on Machine Learning}, volume 202 of \emph{Proceedings of Machine Learning Research}, pp.\  22137--22176. PMLR, 23--29 Jul 2023{\natexlab{a}}.
\newblock URL \url{https://proceedings.mlr.press/v202/liu23am.html}.

\bibitem[Liu et~al.(2023{\natexlab{b}})Liu, Wang, Dao, Zhou, Yuan, Song, Shrivastava, Zhang, Tian, Re, and Chen]{liu2023deja}
Liu, Z., Wang, J., Dao, T., Zhou, T., Yuan, B., Song, Z., Shrivastava, A., Zhang, C., Tian, Y., Re, C., and Chen, B.
\newblock Deja vu: Contextual sparsity for efficient llms at inference time.
\newblock \emph{arXiv preprint arXiv:2310.17157}, 2023{\natexlab{b}}.
\newblock URL \url{https://arxiv.org/abs/2310.17157}.

\bibitem[Ma et~al.(2023)Ma, Fang, and Wang]{maLLMPrunerStructuralPruning2023}
Ma, X., Fang, G., and Wang, X.
\newblock {LLM}-{Pruner}: {On} the {Structural} {Pruning} of {Large} {Language} {Models}, September 2023.
\newblock URL \url{http://arxiv.org/abs/2305.11627}.

\bibitem[Mirzadeh et~al.(2023)Mirzadeh, Alizadeh, Mehta, Del~Mundo, Tuzel, Samei, Rastegari, and Farajtabar]{mirzadeh2023relu}
Mirzadeh, I., Alizadeh, K., Mehta, S., Del~Mundo, C.~C., Tuzel, O., Samei, G., Rastegari, M., and Farajtabar, M.
\newblock Relu strikes back: Exploiting activation sparsity in large language models.
\newblock \emph{arXiv preprint arXiv:2310.04564}, 2023.
\newblock URL \url{https://arxiv.org/abs/2310.04564}.

\bibitem[OpenAI(2023)]{Gpt-4}
OpenAI.
\newblock Openai: Gpt-4, 2023.
\newblock URL \url{https://openai.com/research/gpt-4}.

\bibitem[Paperno et~al.(2016)Paperno, Kruszewski, Lazaridou, Pham, Bernardi, Pezzelle, Baroni, Boleda, and Fern{\'a}ndez]{paperno2016lambada}
Paperno, D., Kruszewski, G., Lazaridou, A., Pham, Q.~N., Bernardi, R., Pezzelle, S., Baroni, M., Boleda, G., and Fern{\'a}ndez, R.
\newblock The lambada dataset: Word prediction requiring a broad discourse context.
\newblock \emph{arXiv preprint arXiv:1606.06031}, 2016.

\bibitem[Qiu et~al.(2024)Qiu, Huang, and Fu]{qiu-etal-2024-unlocking}
Qiu, Z., Huang, Z., and Fu, J.
\newblock Unlocking emergent modularity in large language models.
\newblock In Duh, K., Gomez, H., and Bethard, S. (eds.), \emph{Proceedings of the 2024 Conference of the North American Chapter of the Association for Computational Linguistics: Human Language Technologies (Volume 1: Long Papers)}, pp.\  2638--2660, Mexico City, Mexico, June 2024. Association for Computational Linguistics.
\newblock \doi{10.18653/v1/2024.naacl-long.144}.
\newblock URL \url{https://aclanthology.org/2024.naacl-long.144}.

\bibitem[Riquelme et~al.(2021)Riquelme, Puigcerver, Mustafa, Neumann, Jenatton, Susano~Pinto, Keysers, and Houlsby]{riquelme2021scaling}
Riquelme, C., Puigcerver, J., Mustafa, B., Neumann, M., Jenatton, R., Susano~Pinto, A., Keysers, D., and Houlsby, N.
\newblock Scaling vision with sparse mixture of experts.
\newblock In \emph{Advances in Neural Information Processing Systems}, volume~34, pp.\  8583--8595, 2021.

\bibitem[Sakaguchi et~al.(2020)Sakaguchi, Bras, Bhagavatula, and Choi]{WinoGrande:conf/aaai/SakaguchiBBC20}
Sakaguchi, K., Bras, R.~L., Bhagavatula, C., and Choi, Y.
\newblock Winogrande: An adversarial winograd schema challenge at scale.
\newblock In \emph{The Thirty-Fourth {AAAI} Conference on Artificial Intelligence, {AAAI} 2020, The Thirty-Second Innovative Applications of Artificial Intelligence Conference, {IAAI} 2020, The Tenth {AAAI} Symposium on Educational Advances in Artificial Intelligence, {EAAI} 2020, New York, NY, USA, February 7-12, 2020}, pp.\  8732--8740. {AAAI} Press, 2020.
\newblock \doi{10.1609/AAAI.V34I05.6399}.
\newblock URL \url{https://doi.org/10.1609/aaai.v34i05.6399}.

\bibitem[So et~al.(2022)]{so2022squaredrelu}
So, D. et~al.
\newblock Squared relu: A simple and effective activation function.
\newblock \emph{Advances in Neural Information Processing Systems}, 2022.

\bibitem[Song et~al.(2024{\natexlab{a}})Song, Han, Zhang, Hu, Shi, Li, Chen, Liu, Li, Yang, and Sun]{song2024prosparse}
Song, C., Han, X., Zhang, Z., Hu, S., Shi, X., Li, K., Chen, C., Liu, Z., Li, G., Yang, T., and Sun, M.
\newblock Prosparse: Introducing and enhancing intrinsic activation sparsity within large language models.
\newblock \emph{arXiv preprint arXiv:2402.13516}, 2024{\natexlab{a}}.
\newblock URL \url{https://arxiv.org/abs/2402.13516}.

\bibitem[Song et~al.(2023)Song, Mi, Xie, and Chen]{song2023powerinfer}
Song, Y., Mi, Z., Xie, H., and Chen, H.
\newblock Powerinfer: Fast large language model serving with a consumer-grade gpu.
\newblock \emph{arXiv preprint arXiv:2312.12456}, 2023.
\newblock URL \url{https://arxiv.org/abs/2312.12456}.

\bibitem[Song et~al.(2024{\natexlab{b}})]{song2024activation}
Song, Y. et~al.
\newblock Powerinfer: Enhancing activation sparsity in llm serving with consumer-grade gpus.
\newblock \emph{arXiv preprint arXiv:2312.12456}, 2024{\natexlab{b}}.
\newblock URL \url{https://arxiv.org/abs/2312.12456}.

\bibitem[Team(2021)]{mosaicml2022composer}
Team, T. M.~M.
\newblock composer.
\newblock \url{https://github.com/mosaicml/composer/}, 2021.

\bibitem[TogetherAI(2023)]{Redpajama}
TogetherAI.
\newblock Redpajama: An open source recipe to reproduce llama training dataset, 2023.

\bibitem[Touvron et~al.(2023{\natexlab{a}})Touvron, Martin, Stone, Albert, Almahairi, Babaei, Bashlykov, Batra, Bhargava, Bhosale, Bikel, Blecher, Ferrer, Chen, Cucurull, Esiobu, Fernandes, Fu, Fu, Fuller, Gao, Goswami, Goyal, Hartshorn, Hosseini, Hou, Inan, Kardas, Kerkez, Khabsa, Kloumann, Korenev, Koura, Lachaux, Lavril, Lee, Liskovich, Lu, Mao, Martinet, Mihaylov, Mishra, Molybog, Nie, Poulton, Reizenstein, Rungta, Saladi, Schelten, Silva, Smith, Subramanian, Tan, Tang, Taylor, Williams, Kuan, Xu, Yan, Zarov, Zhang, Fan, Kambadur, Narang, Rodriguez, Stojnic, Edunov, and Scialom]{touvronLlamaOpenFoundation2023}
Touvron, H., Martin, L., Stone, K., Albert, P., Almahairi, A., Babaei, Y., Bashlykov, N., Batra, S., Bhargava, P., Bhosale, S., Bikel, D., Blecher, L., Ferrer, C.~C., Chen, M., Cucurull, G., Esiobu, D., Fernandes, J., Fu, J., Fu, W., Fuller, B., Gao, C., Goswami, V., Goyal, N., Hartshorn, A., Hosseini, S., Hou, R., Inan, H., Kardas, M., Kerkez, V., Khabsa, M., Kloumann, I., Korenev, A., Koura, P.~S., Lachaux, M.-A., Lavril, T., Lee, J., Liskovich, D., Lu, Y., Mao, Y., Martinet, X., Mihaylov, T., Mishra, P., Molybog, I., Nie, Y., Poulton, A., Reizenstein, J., Rungta, R., Saladi, K., Schelten, A., Silva, R., Smith, E.~M., Subramanian, R., Tan, X.~E., Tang, B., Taylor, R., Williams, A., Kuan, J.~X., Xu, P., Yan, Z., Zarov, I., Zhang, Y., Fan, A., Kambadur, M., Narang, S., Rodriguez, A., Stojnic, R., Edunov, S., and Scialom, T.
\newblock Llama 2: {Open} {Foundation} and {Fine}-{Tuned} {Chat} {Models}, July 2023{\natexlab{a}}.
\newblock URL \url{http://arxiv.org/abs/2307.09288}.

\bibitem[Touvron et~al.(2023{\natexlab{b}})Touvron, Martin, Stone, Albert, Almahairi, Babaei, Bashlykov, Batra, Bhargava, Bhosale, et~al.]{touvron2023llama}
Touvron, H., Martin, L., Stone, K., Albert, P., Almahairi, A., Babaei, Y., Bashlykov, N., Batra, S., Bhargava, P., Bhosale, S., et~al.
\newblock Llama 2: Open foundation and fine-tuned chat models.
\newblock \emph{arXiv preprint arXiv:2307.09288}, 2023{\natexlab{b}}.

\bibitem[Vaswani et~al.(2023)Vaswani, Shazeer, Parmar, Uszkoreit, Jones, Gomez, Kaiser, and Polosukhin]{vaswani2023attentionneed}
Vaswani, A., Shazeer, N., Parmar, N., Uszkoreit, J., Jones, L., Gomez, A.~N., Kaiser, L., and Polosukhin, I.
\newblock Attention is all you need, 2023.
\newblock URL \url{https://arxiv.org/abs/1706.03762}.

\bibitem[Wang et~al.(2024)Wang, Ma, Cao, Zhang, Xue, Shi, Zheng, Miao, Yang, Cao, Yang, and Yang]{wang2024ladder}
Wang, L., Ma, L., Cao, S., Zhang, Q., Xue, J., Shi, Y., Zheng, N., Miao, Z., Yang, F., Cao, T., Yang, Y., and Yang, M.
\newblock Ladder: Enabling efficient low-precision deep learning computing through hardware-aware tensor transformation.
\newblock In \emph{18th USENIX Symposium on Operating Systems Design and Implementation (OSDI 24)}, 2024.
\newblock URL \url{https://www.usenix.org/conference/osdi24/presentation/wang-lei}.

\bibitem[Welbl et~al.(2017)Welbl, Liu, and Gardner]{sciqa}
Welbl, J., Liu, N.~F., and Gardner, M.
\newblock Crowdsourcing multiple choice science questions.
\newblock \emph{arXiv preprint arXiv:1707.06209}, 2017.

\bibitem[Wu et~al.(2024)Wu, Huang, and Wei]{wu2024multihead}
Wu, X., Huang, S., and Wei, F.
\newblock Multi-head mixture-of-experts.
\newblock \emph{arXiv preprint arXiv:2404.15045}, 2024.
\newblock URL \url{https://arxiv.org/abs/2404.15045}.

\bibitem[Xia et~al.(2023)Xia, Gao, Zeng, and Chen]{xiaShearedLLaMAAccelerating2023}
Xia, M., Gao, T., Zeng, Z., and Chen, D.
\newblock Sheared {LLaMA}: {Accelerating} {Language} {Model} {Pre}-training via {Structured} {Pruning}, October 2023.
\newblock URL \url{http://arxiv.org/abs/2310.06694}.

\bibitem[Xue et~al.(2024{\natexlab{a}})Xue, Zheng, Fu, Ni, Zheng, Zhou, and You]{xue2024openmoeearlyeffortopen}
Xue, F., Zheng, Z., Fu, Y., Ni, J., Zheng, Z., Zhou, W., and You, Y.
\newblock Openmoe: An early effort on open mixture-of-experts language models, 2024{\natexlab{a}}.
\newblock URL \url{https://arxiv.org/abs/2402.01739}.

\bibitem[Xue et~al.(2024{\natexlab{b}})Xue, Zheng, Fu, Ni, and Zhou]{xue2024openmoe}
Xue, F., Zheng, Z., Fu, Y., Ni, J., and Zhou, W.
\newblock Openmoe: An early effort on open mixture-of-experts language models.
\newblock \emph{arXiv preprint arXiv:2402.01739}, 2024{\natexlab{b}}.
\newblock URL \url{https://arxiv.org/abs/2402.01739}.

\bibitem[You et~al.(2022)You, Bhojanapalli, Li, and Rawat]{you2022lazyneuron}
You, C., Bhojanapalli, S., Li, D., and Rawat, A.
\newblock The lazy neuron phenomenon: On emergence of activation sparsity in transformers.
\newblock \emph{arXiv preprint arXiv:2210.06313}, 2022.
\newblock URL \url{https://arxiv.org/abs/2210.06313}.

\bibitem[Zellers et~al.(2019)Zellers, Holtzman, Bisk, Farhadi, and Choi]{HellaSwag:conf/acl/ZellersHBFC19}
Zellers, R., Holtzman, A., Bisk, Y., Farhadi, A., and Choi, Y.
\newblock Hellaswag: Can a machine really finish your sentence?
\newblock In Korhonen, A., Traum, D.~R., and M{\`{a}}rquez, L. (eds.), \emph{Proceedings of the 57th Conference of the Association for Computational Linguistics, {ACL} 2019, Florence, Italy, July 28- August 2, 2019, Volume 1: Long Papers}, pp.\  4791--4800. Association for Computational Linguistics, 2019.
\newblock \doi{10.18653/V1/P19-1472}.
\newblock URL \url{https://doi.org/10.18653/v1/p19-1472}.

\bibitem[Zhang et~al.(2024)Zhang, Song, Yu, Han, Lin, Xiao, Song, Liu, Mi, and Sun]{zhang2024relu2}
Zhang, Z., Song, Y., Yu, G., Han, X., Lin, Y., Xiao, C., Song, C., Liu, Z., Mi, Z., and Sun, M.
\newblock Relu2 wins: Discovering efficient activation functions for sparse llms.
\newblock \emph{arXiv preprint arXiv:2402.03804}, 2024.
\newblock URL \url{https://arxiv.org/abs/2402.03804}.

\bibitem[Zhou et~al.(2022)Zhou, Lei, Liu, Du, Huang, Zhao, Dai, Le, Laudon, et~al.]{zhou2022mixture}
Zhou, Y., Lei, T., Liu, H., Du, N., Huang, Y., Zhao, V., Dai, A.~M., Le, Q.~V., Laudon, J., et~al.
\newblock Mixture-of-experts with expert choice routing.
\newblock In \emph{Advances in Neural Information Processing Systems}, 2022.

\end{thebibliography}
